\documentclass[11pt]{article}
\usepackage[utf8]{inputenc}
\usepackage{amsmath}
\usepackage{wrapfig}
\usepackage{amssymb}
\usepackage{mathrsfs}
\usepackage{color}
\usepackage{graphicx}
\usepackage{mathabx}
\usepackage{dsfont}
\usepackage{float}
\usepackage{epsfig}
\usepackage{lscape}
\usepackage[bottom]{footmisc}
\usepackage{multirow}
\usepackage{caption}
\usepackage{url}
\usepackage{stmaryrd}
\usepackage{booktabs}
\usepackage{todonotes}

\usepackage{rotating}
\usepackage{epstopdf}

\newtheorem{theorem}{Theorem}[section]

\newtheorem{proposition}{Proposition}[section]
\newtheorem{corollary}{Corollary}[section]
\newtheorem{lemma}{Lemma}[section]
\newtheorem{remark}{Remark}[section]

\makeatletter

\@addtoreset{equation}{section}
\makeatother

\def\ind{ {{\rm 1}\hskip-2.2pt{\rm l}}}

\begin{document}

\title{Survival Estimation for Missing not at Random Censoring Indicators based on Copula Models}

\author{
{\large Mikael Escobar-Bach}
\footnote{Email: \texttt{mikael.escobar-bach@univ-angers.fr}.  \vspace*{.1cm}} \\ \textit{LAREMA, Université d'Angers} 
\and 
{\large Olivier Goudet} \footnote{Email: \texttt{olivier.goudet@univ-angers.fr}}\\ \textit{LERIA, Université d'Angers} 
}
\maketitle

\begin{abstract}
In the presence of right-censored data with covariates, the conditional Kaplan-Meier estimator (also known as the Beran estimator) consistently estimates the conditional survival function of the random follow-up for the event of interest. However, a necessary condition is the unambiguous knowledge of whether each individual is censored or not, which may be incomplete in practice. We therefore propose a study of the Beran estimator when the censoring indicators are generic random variables and discuss necessary conditions for the efficiency of the Beran estimator. From this, we provide a new estimator for the conditional survival function with missing not at random (MNAR) censoring indicators based on a conditional copula model for the missingness mechanism. In addition to the theoretical results, we illustrate how the estimators work for small samples through a simulation study and show their practical applicability by analyzing synthetic and real data.
\end{abstract}



\maketitle

\section{Introduction}

As the volume of data increases, the problem of missing data has become increasingly present in many areas of statistical applications. In the classical literature of survival analysis, the study of the duration time preceding an event of interest is considered with a series of random censors that may prevent the entire survival period from being captured. This is known as the censoring mechanism and arises from limitations depending on the nature of the study. For example, this feature is particularly present in medicine, with studies of survival times to recovery/decease from a particular chronic disease. In fact, a patient's lifetime or time to occurrence is not observed if it exceeds the study follow-up time, or because of the early withdrawal of the individual from the study population. This is referred to as a right-censored observation, indicating that the event of interest may only occur after this time. Other examples of right censoring can be found in a wide range of fields, such as economics (duration of unemployment), social science (time before marriage or childbearing), or actuarial science (life insurance or time before claims).

In survival analysis, it is often of interest to estimate the survival function of the random follow-up time of the event of interest. The most popular and well-known non-parametric approach is a product-limit estimator, so-called the Kaplan-Meier estimator, firstly introduced in \cite{Kaplan1958}. The latter has been extensively studied in the last decades and shows appealing properties; their asymptotic properties were proved in \cite{Foldes1981,Wellner1982} and discussions on the limit law process were proposed in \cite{Deheuvels2000,Ouadah2013}. It is also worth mentioning that its convergences rates have been widely studied, among others in \cite{Horvath1983,Chen1997} or \cite{Wellner2008} with exponential bounds for the empirical process. In cases where the data come along with the response of random covariates, one can consider estimating the conditional survival function using the Beran estimator \cite{Beran1981}, which is a direct extension of the Kaplan-Meier estimator through a kernel method to the regression context. Naturally, the Beran estimator has inherited the same interests as its non-conditional counterpart and has been intensively studied by \cite{Dabrowska1987,Dabrowska1989,Dabrowska1992} or \cite{Li1995} among others. In parallel, several studies have proposed various adaptations of the Kaplan-Meier estimator by replacing the censoring indicators with other estimates. Historically, this idea first appears independently in \cite{Abdushukurov1987} and \cite{Cheng1987} where $\delta$ is replaced by its conditional expectation when equals to 1 \cite{Dikta1998}. From another perspective, \cite{Wang2008}, \cite{Subramanian2011} and \cite{Brunel2014} proposed similar consistent estimators for the missing censoring indicators model, where in this context, a non-empty fraction of the censoring indicators are missing completely (MCAR) or simply (MAR) at random. The MCAR and MAR mechanisms have been widely studied in the literature and occur when there is independence between the outcomes and the missing pattern: for MCAR, the missing mechanism and the outcome are independent, while for MAR, the missingness depends only on the observed data components, and not on the missing components. When neither the MCAR nor the MAR mechanisms are valid, i.e. the missingness and the unobserved data are dependent, we say that the mechanism is missing not at random (MNAR). In this latter case, the unobserved data cannot be directly identified without a specified joint model between the outcomes and the missing mechanism, which renders any statistical approach to the MCAR and/or MAR contexts non-valid \cite{Little2002}.\\ 

Here, we propose to handle the problem of missing censoring indicators in the MNAR scenario when our missingness mechanism is described by a copula model with a known dependence structure. A particular example with a gaussian structure was introduced in \cite{Heckman1976,Heckman1979} via the Heckman's model for continuous variables where the outcome and the missingness attribute follow a linear regression model linked by their error terms. In the proposed version, we assume some known but arbitrary dependence structure between the censoring and the missing indicators. Although we require the full knowledge for the copula function, we impose mild conditions on the marginal distributions, since they only need to belong to classes of functions with finite \textit{Vapnik Chervonenkis} dimension. This allows the approximation of complex dependency patterns with covariates without assuming specific knowledge of the underlying individual generative process. In order to estimate all the parameters of the joint bivariate model, we introduce a two-fold maximum likelihood approach based on \cite{galimard2018heckman}. An end-to-end learning framework is then introduced to simultaneously learn the outcome model, the selection model and the dependency model by gradient descent. As such, we particularly consider the class of neural networks. Using neural networks for time-to-event models has already been done in the past, but never in the MNAR setting. Some existing works in this direction focus on extending the Cox model with non-linear models of the hazard rate. This type of work dates back to \cite{faraggi1995neural} but has recently been revisited with modern deep learning techniques \cite{katzman2016deep} and with convolutional neural networks to build a time-to-event Cox model from images \cite{zhu2016deep}. From another perspective, recent deep learning models do not rely on the Cox formulation and directly learn the estimated joint distribution of survival time without making assumptions on the relationship between covariates and hazard rate \cite{lee2018deephit}, or leverage on adversarial learning  for non-parametric estimation of time-to-event distributions \cite{chapfuwa2018adversarial}.\\

In the first part, we thus propose to study the asymptotic behavior for the Beran estimator whenever the censoring indicators are drawn from any generic random variable $P\in[0,1]$ and derive its almost sure representation under mild conditions. This allows us to show the estimator consistency and optimality in terms of asymptotic variance as long as $P$ shares the sames conditional expectation of $\delta$. For imputation methods over the censoring indicators, we propose necessary conditions to ensure that the Beran estimator keeps the same variance at the asymptotic. In particular, we obtain the functional convergence for the empirical process when an estimator of the conditional expectation of $\delta$ assumes proper convergence rates. Following the theoretical analysis, we secondly consider a plug-in estimate of the survival function based on the outcomes of a generalized Heckman's model and show that it fits the aforementioned theoretical guarantees, along with a short simulation study with alternative strategies for the estimation of the conditional expectation of $\delta$ in the MAR and MNAR settings. The rest of the paper is organized as follows. The framework for the i.i.d. censoring model with arbitrary censoring indicators, which is a generalization of the survival analysis model, is presented in subsection \ref{section_model}. In the subsection \ref{section_cdf}, we establish the almost-sure representation for our generalized Beran estimator and provide sufficient assumptions to ensure the weak convergence of the empirical process towards a mean-zero Gaussian process. The weak convergence for the plug-in estimators is guaranteed for appropriate convergence rates and is discussed in Section \ref{sec:plugin}. Comparative and numerical results are displayed in Section \ref{sec:expe} and a real application with data from patients with prostate cancer is presented in Section \ref{sec:expe_real}. The proofs are postponed in the supplementary material.

\section{Generalized Censoring Indicator}

In this section, we initially study the asymptotic behavior for the Beran estimator whenever the censoring indicator is a general random variable. This will be useful in the sequel when the missing censoring indicators will be replaced by some estimation for the conditional censoring probability.

\subsection{Model and Estimators}
\label{section_model}

\noindent
We consider a random vector $(Y,C,P,X)\in\mathbb{R}\times\mathbb{R}\times[0,1]\times\mathbb{R}^p$ under a random right-censoring model, in the sense that the data at hands are restricted to $(T,P,X)$ where $T:=\min(Y,C)$, $P$ is the general censorship indicator and $X$ is an explanatory random covariable with density function $f$. Note that we will also consider the classic censorship model given by $P=\delta:=\ind_{\{Y\leq C\}}$ where $\delta$ is the common censoring indicator. The conditional distribution functions of the survival and the censoring times are respectively denoted by $F$ and $G$.  Furthermore, we adopt the notation $S_Z$ to define the support of any random variable $Z$. Recall also that the right end points of the support of the distributions $F$ and $G$ are respectively denoted by $\tau_F(x)$ and $\tau_G(x)$. Finally, we denote $F^\leftarrow$ the generalized inverse function for the distribution function $F$ given by
\begin{eqnarray*}
F^\leftarrow(\alpha|x):=\inf\{t,F(t|x)\geq \alpha\},\quad\alpha\in(0,1).
\end{eqnarray*}
We will work under minimal conditions on the distribution functions, although we have to impose the usual identification assumption with non-informative censoring, in the sense that $Y$ and $C$ are independent. This implies that the distribution of the observation time $H(t|x):=\mathbb{P}(T \le t|X=x)$ satisfies $1-H(t|x)=(1-F(t|x))(1-G(t|x))$. Here, the conditional right-end point of $H$ is denoted by $\tau_H(x)=\min(\tau_F(x),\tau_G(x))$. In the sequel, we will also use the notations
\begin{eqnarray*}
H^u(t|x)&:=&\mathbb{P}(T\leq t,\delta=1|X=x) =\int_{-\infty}^t (1-G(s^-|x)) dF(s|x)\\
\text{and}\quad H_p^u(t|x)&:=&\mathbb{E}[P\ind_{\{T\leq t\}}|X=x],
\end{eqnarray*}
for the sub-distribution functions of the uncensored observations and $\Lambda(\cdot|x)$ for the cumulative hazard function given by
\begin{eqnarray*} 
\Lambda(t|x):=\int_{-\infty}^t \dfrac{dH^u(s|x)}{1-H(s^-|x)}
\end{eqnarray*} 
for any $t\in\mathbb{R}$. We next define the Beran estimator for the conditional distribution $F$. Let assume an independent and identically distributed (\textit{i.i.d.}) $n$-sized sample drawn from the classical censoring model $\{(T_i,\delta_i,X_i)\}_{1\leq i\leq n}$. Denote the $i$-th order statistic of $T_1,\ldots,T_n$ by $T_{(i)}$ and its corresponding censoring indicator and covariable by $\delta_{(i)}$ and $X_{(i)}$ respectively. In the absence of ties, the Beran estimator is given by
\begin{eqnarray*}
F_n(t|x):=1-\prod_{T_{(i)}\leq t,\,\delta_{(i)}=1}\left(1-\dfrac{W_b(x-X_{(i)})}{1-\sum_{j=1}^{i-1}W_b(x-X_{(j)})}\right),
\end{eqnarray*}
where for any $i=1,\ldots,n$,
\begin{eqnarray*}
W_b(x-X_i):=\dfrac{K_b(x-X_i)}{\sum_{j=1}^nK_b(x-X_j)}
\end{eqnarray*}
and $K_b(\cdot)=K(\cdot/b)/b^p$ with $K$ a kernel function and $b=b_n$ a non-random positive sequence such that $b_n\rightarrow 0$ as $n\rightarrow \infty$. Adapted to our context, we assume from now on that we have at our disposal another $n$-sized \textit{i.d.d.} sample $\{(T_i,P_i,X_i)\}_{1\leq i\leq n}$ but drawn the triplet $(T,P,X)$. The new estimator of the distribution function $F$ is similarly defined as 
\begin{eqnarray*}
\widehat{F}_n(t|x):=1-\prod_{T_{(i)}\leq t}\left(1-\dfrac{W_b(x-X_{(i)})}{1-\sum_{j=1}^{i-1}W_b(x-X_{(j)})}\right)^{P_{(i)}}
\end{eqnarray*}
where the product over an empty set is also defined to be 1. It is worth noting that the times of the jumps are the same for $F_n$ and $\widehat{F}_n$, but the jump sizes are different, especially when $\delta_i=0$ with no jump at $T_i$ for $F_n$. In order to further study $\widehat{F}_n$, we need to introduce the empirical estimators for the distribution function $H$, $H^u$, $H^u_p$ and $\Lambda$ respectively given by
\begin{gather*}
H_n(t|x):=\sum_{i=1}^nW_b(x-X_i)\ind_{\{T_i\leq t\}},\quad
H_n^u(t|x):=\sum_{i=1}^nW_b(x-X_i)\ind_{\{T_i\leq t,\delta_i=1\}},\\ \widehat{H}_n^u(t|x):=\sum_{i=1}^nW_b(x-X_i)\ind_{\{T_i\leq t\}}P_i
\end{gather*}
and
\begin{eqnarray*}
\Lambda_n(t|x) := \int_{-\infty}^t\dfrac{dH_n^u(s|x)}{1-H_n(s^-|x)},\quad \widehat{\Lambda}_n(t|x) := \int_{-\infty}^t\dfrac{d\widehat{H}_n^u(s|x)}{1-H_n(s^-|x)}.
\end{eqnarray*}
We finally denote $f_n$ the the kernel estimator for the density function $f$ with $\forall x\in\mathbb{R}^p$
\begin{eqnarray*}
f_n(x):=\dfrac{1}{n}\sum_{i=1}^nK_b(x-X_i).
\end{eqnarray*}

\bigskip
\subsection{Asymptotic Properties}
\label{section_cdf}

In this section, we derive and discuss the asymptotic properties of the estimator $\widehat{F}_n$. We also assume from now on that $x\in\text{int}(S_X)$ defines a fixed reference position such that $f(x)>0$. Due to the regression context, we need some H\"older-type conditions on the distribution functions $H$, $H^u$, $H^u_p$, $F$ and on the density function $f$. Let $\Vert \cdot \Vert$ be any norm on $\mathbb{R}^p$.\\

\noindent
{\bf Assumption $(\mathcal{H})$.} There exist $0<\eta,\eta' \le 1$ and $c>0$ such that for any $t,s \in\mathbb{R}$ and any $x_1,x_2\in S_X$,
\begin{eqnarray*}
&(\mathcal{H}.1)&\hspace{.2cm}|f(x_1)-f(x_2)|\leq c\Vert x_1-x_2\Vert^\eta\\
&(\mathcal{H}.2)&\hspace{.2cm}|H(t|x_1)-H(t|x_2)|\leq c\Vert x_1-x_2\Vert^\eta\\
&(\mathcal{H}.3)&\hspace{.2cm}|H^u(t|x_1)-H^u(s|x_2)|\leq c(\Vert x_1-x_2\Vert^\eta+|t-s|^{\eta'})\\
&(\mathcal{H}.4)&\hspace{.2cm}|H^u_p(t|x_1)-H^u_p(s|x_2)|\leq c(\Vert x_1-x_2\Vert^\eta+|t-s|^{\eta'})\\
&(\mathcal{H}.5)&\hspace{.2cm}|F(t|x)-F(s|x)|\leq c|t-s|^{\eta'}
\end{eqnarray*}
Also, some common assumptions on the kernel function as well as the continuity of $G$ need to be imposed.\\

\noindent
{\bf Assumption $(\mathcal{K})$.} Let $K$ be a bounded density function in $\mathbb R^p$ with support $S_K$ included in the unit ball of $\mathbb{R}^p$ with respect to norm $\Vert \cdot \Vert$.\\ 

\noindent
{\bf Assumption $(\mathcal{G})$.} The distribution function $G(.|x)$ is continuous.\\

\noindent
It is usually difficult to work with results for empirical processes on product type estimators, likewise the Beran estimator. It hence common to consider their logarithm transformation over a compact support. Due to those technical reasons, we study the asymptotic behavior of our estimator through $\widehat{\Lambda}_n$ using the approximation $1-\widehat{F}_n\approx\exp(-\widehat{\Lambda}_n)$ as given in the following lemma.
\begin{lemma}
\label{lambda}
Let $\tau_0<\tau_1<\tau_H(x)$. Then we have that
\begin{eqnarray*}
\sup_{t\in[\tau_0,\tau_1]}\left|1-\widehat{F}_n(t|x)-\exp(-\widehat{\Lambda}_n(t|x))\right| &\leq& \Vert K\Vert_\infty \dfrac{\widehat{H}_n^u(T|x)}{nb^pf_n(x)(1-H_n(\tau_1|x))^2}\\
&=&\mathcal{O}_\mathbb{P}\left((nb^p)^{-1}\right).
\end{eqnarray*}
In particular, it turns out that
\begin{eqnarray*}
\sup_{t\in[\tau_0,\tau_1]}\sqrt{nb^p}\left|1-\widehat{F}_n(t|x)-\exp(-\widehat{\Lambda}_n(t|x))\right|=o_\mathbb{P}(1)
\end{eqnarray*}  
when $f(x)>0$. 
\end{lemma}
This results ensures that the asymptotic properties of $\widehat{\Lambda}_n(.|x)$ are equivalent to than that of $\widehat{F}_n(.|x)$ on $[\tau_0,\tau_1]$. We hereby prove the almost-sure representation for the estimator of our generalized cumulative hazard function.
\begin{proposition}\label{prop:asrep}
Under the assumptions $(\mathcal{H})$, $(\mathcal{K})$ and $(\mathcal{G})$, for any $\tau_0<\tau_1<\tau_H(x)$, we have for $\tau_0\leq t\leq \tau_1$ and $nb^{2\eta+p}|\log b|=\mathcal{O}(1)$
\begin{eqnarray*}
\widehat{\Lambda}_n(t|x)-\Lambda(t|x)=\sum_{i=1}^nW_b(x-X_i)\widehat{\ell}(t,T_i,\delta_i,P_i|x)+r_n(t|x)
\end{eqnarray*}
where
\begin{eqnarray}
\label{eq:ell}
\widehat{\ell}(t,T_i,\delta_i,P_i|x)&=& \dfrac{\ind_{\{T_i \leq t,\delta_i=1\}}-H^u(t|x)}{1-H(t|x)} -\int_{-\infty}^t\dfrac{\ind_{\{T_i \leq s,\delta_i=1\}}-H^u(s|x)}{(1-H(s|x))^2}dH(s|x)\\
&& \nonumber +\int_{-\infty}^t\dfrac{\ind_{\{T_i < s\}}-H(s|x)}{(1-H(s|x))^2}dH_p^u(s|x) +\dfrac{(P_i-\delta_i)\ind_{\{T_i\leq t\}}}{1-H(T_i|x)}
\end{eqnarray}
and
\begin{eqnarray*}
\sup_{\tau_0\leq t\leq \tau_1}|r_n(t|x)|=\mathcal{O}_\mathbb{P}((nb^p)^{-3/4} |\log b|^{3/4}).
\end{eqnarray*}
\end{proposition}
 As expected, one can retrieve the usual almost-sure representation for $\Lambda_n$ when $P=\delta$. It is however worth mentioning that we can distinguish two parts in the definition of $\widehat{\ell}$. On one hand, the elements on the first line in (\ref{eq:ell}) are unbiased with no dependency with regards to $P$. On the other hand, the $\{P_i-\delta_i\}_{1\leq i\leq n}$ in the second line are the only elements that could result in a biased part for $\widehat{\Lambda}_n$. The condition 
\begin{eqnarray*}
\mathbb{E}\left[\left.\dfrac{(P-\delta)\ind_{\{T\leqslant t\}}}{1-H(T|x)}\right| X=x\right]=0,\quad\tau_0\leqslant t\leqslant\tau_1
\end{eqnarray*}
hence appears sufficient to ensure that $\widehat{\Lambda}_n$, and subsequently $\widehat{F}_n$, are unbiased estimators. We thus introduce the following equivalent assumption\\

\noindent
\textbf{Assumption ($\mathcal{E}$)} : Let $S_T$ stands for the support of $T$, then
\begin{eqnarray}
\label{model}
\forall t\in S_T\subset\mathbb{R},\quad\mathbb{E}[P|T=t,X=x]=\mathbb{P}(Y\leq C|T=t,X=x).
\end{eqnarray}
Combining the assumption below together with the proposition \ref{prop:asrep}, we are able to establish an almost-sure representation result for $\widehat{F}_n-F$ likewise $\widehat{\Lambda}_n-\Lambda$.
\begin{corollary}
\label{corollary_Fn}
Under the assumptions of Proposition \ref{prop:asrep}, if the assumption ($\mathcal{E}$) is verified, then
\begin{eqnarray*}
\widehat{F}_n(t|x)-F(t|x)&=&(1-F(t|x))\sum_{i=1}^nW_b(x-X_i)\widehat{\ell}(t,T_i,\delta_i,P_i|x) + r_n(t|x)
\end{eqnarray*}
where 
\begin{eqnarray*}
\sup_{\tau_0\leq t\leq \tau_1}|r_n(t|x)|=\mathcal{O}_\mathbb{P}((nb^p)^{-3/4} |\log b|^{3/4}).
\end{eqnarray*}
\end{corollary}
This finally allows us to obtain the main result of this subsection, which is the weak convergence of the estimator $\widehat{F}_n$ as a process in $\mathcal{L}^\infty[\tau_0,\tau_1]$ for any $\tau_0<\tau_1 < \tau_H(x)$ and for fixed $x\in\mathcal{S}_X$.  Here, for any set $S$ we define the space $\mathcal{L}^\infty(S)$ as the space of bounded functions with values on $S$ and endowed with the uniform norm.
\begin{theorem}
\label{theorem_nesti}
Assume $(\mathcal{E})$, $(\mathcal{H})$, $(\mathcal{K})$, $(\mathcal{G})$, and assume that $f(x)>0$, $nb^p |\log b|^{-3} \to \infty$ and $nb^{2\eta+p-q} |\log b|^{-1} =\mathcal{O}(1)$ for some $q>0$. Then, for any $\tau_0 < \tau_1<\tau_H(x)$, the process
\begin{eqnarray}
\label{process}
\left\{(nh^p)^{1/2}(\widehat{F}_n(t|x)-F(t|x)),\quad t\in [\tau_0,\tau_1]\right\},
\end{eqnarray}
converges weakly in $\mathcal{L}^\infty[\tau_0,\tau_1]$ to a continuous mean-zero Gaussian process $Z(\cdot|x)$ with covariance structure
\begin{eqnarray*}
\widehat{\Gamma}(t,s|x)=\dfrac{\Vert K\Vert_2^2}{f(x)}(1-F(t|x))(1-F(s|x)) \int_{-\infty}^{t\wedge s}\dfrac{\mathbb{E}[P^2|T=y, X=x] }{(1-H(y|x))^2}dH(y|x).\\
\end{eqnarray*}
In particular, $\widehat{\Gamma}(t,s|x)$ is minimal for $P=\mathbb{P}(Y\leqslant C|T,X)$ a.s. 
\end{theorem}
It turns out that $\widehat{F}_n$ is a consistent estimator for any random censorship indicator $P$ when the assumption $(\mathcal{E})$ is verified. This particularly makes the random variable $p(T,X):=\mathbb{P}(Y\leqslant C|T,X)$ a natural candidate since it satisfies the assumption ($\mathcal{E}$) and provides a minimal asymptotic variance for $\widehat{F}_n$. However, the function $p$ is usually unknown in practice, which explains the interest in studying the behavior of $\widehat{F}_n$ when $P$ is replaced by a proper estimator of $p$. We hereafter will consider various versions of estimators of the function $p$ with the common notation $p_n$. All along, we define the generalized Beran estimator (GBE) as a plug-in estimate for $F$ given by
\begin{eqnarray}
\widecheck{F}_n(t|x)=1-\prod_{T_{(i)}\leq t}\left(1-\dfrac{W_b(x-X_{(i)})}{1-\sum_{j=1}^{i-1}W_b(x-X_{(j)})}\right)^{p_n(T_i,X_i)}
\label{eq:GBE_estimator}
\end{eqnarray}
whereas $\widehat{F}_n$ will be associated to censoring indicator $P=p(T,X)$ in the rest of the paper. In the next proposition, we ensure that $\widecheck{F}_n$ inherits the same asymptotic properties than $\widehat{F}_n$ if $p_n$ features a fast enough $L_q$ convergence rate when converging towards $p$. 
\begin{proposition}
\label{prop:plugin}
Let $p_n$ be any proper estimator of $p$ in the sense that $\exists\, q>1$ such that
\begin{eqnarray}
\label{assumption_pn}
\mathbb{E}\left[\left|p_n(T,X)-p(T,X)\right|^q\right]^{1/q}=o((nb^p)^{-1/2}).
\end{eqnarray}
Then, it similarly follows that
\begin{eqnarray*}
\sup_{t\in[\tau_0,\tau_1]}\left|\widecheck{F}_n(t|x)-\widehat{F}_n(t|x)\right|=o_\mathbb{P}((nb^p)^{-1/2}).
\end{eqnarray*}
\end{proposition}

\begin{remark}
Although the work in this section will be helpful in the context of missing censoring indicator, the result of the Theorem \ref{theorem_nesti} is also interesting in a more common scenario. The derivation of the asymptotic covariance structure actually shows that it always of interest to replace the common censoring indicators $\delta$ by a proper estimator of the function $p$ applied at $(T,X)$, in the sense of the Proposition \ref{prop:plugin}. This would first ensure a minimal asymptotic variance for $\widecheck{F}_n$, plus shows smoother curves than $F_n$ in practice since the jumps at the censored observation times are certainly not zero. 
\end{remark}
\bigskip
\section{Missing Censoring Indicators}
\label{sec:plugin}
In the sequel, it became clear that the censoring indicators play an important role in the estimation of the survival distribution. When missing, it is hence difficult to expect that the survival estimators remain unbiased if one simply removes the missing data. For the sake of clarity, we consider that the missingness pattern is described through a random binary outcome $\xi$, such that the $i$-th individual's indicator is observed when $\xi_i=1$, and missing when $\xi_i=0$. One approach, commonly referred to as the inverse probability weighting (IPW) method, consists of weighting the available outcomes by the inverse proportion of the observed individual. In our case, this is equivalent to considering the general indicators for $i=1,\ldots,n$
\begin{eqnarray*}
P^{ipw}_i := \frac{\xi_i\delta_i}{q_n(T_i,X_i)} \quad \text{with}\quad q_n(t,x):=\sum_{i=1}^n\xi_iW_{\bar{b}}(x-X_i)\dfrac{\widetilde{K}_{\bar{b}}(t-T_i)}{\sum_{j=1}^n\widetilde{K}_{\bar{b}}(t-T_j)}.
\end{eqnarray*}
where $\widetilde{K}_{\bar{b}}(\cdot)=\widetilde{K}(\cdot/\bar{b})/\bar{b}$ and $\widetilde{K}$ is another kernel function and $\bar{b}$ another bandwidth. This approach is asymptotically similar to \cite{Subramanian2003} with covariables. It however fails to always guarantee unbiased survival estimators, unless we have the independence between the missingness mechanism and the survival data, i.e. under the MAR scenario. The next lemma resumes this property and is a direct consequence of Proposition \ref{prop:asrep}. 
\begin{lemma}
Assume that the random indicator $\xi$ is independent of the $(T,\delta)$ conditionally on $X$. Under the assumptions of Proposition \ref{prop:asrep}, the statistic
\begin{eqnarray*}
\sum_{i=1}^nW_b(X_i-x)\dfrac{(P^{ipw}_i-\delta_i)\ind_{\{T_i\leqslant t\}}}{1-H(T_i|x)}
\end{eqnarray*}
is asymptotically unbiased and so are the estimators $\widehat{\Lambda}_n$ and $\widehat{F}_n$. 
\end{lemma}
 
 In the less restrictive case of MNAR censoring indicators, it is then important to consider the dependence structure driving the joint law of $(\delta,\xi)$ conditionally on $(T,X)$. Following Sklar's theorem \cite{Sklar1959}, the copula functions allow a free margins description of the dependence structure for any random vector where the copula $D$ takes the form of a multivariate distribution function with uniform margins.
 Since we deal with binary random variables, this is equivalent to assuming that $(\delta,\xi)$ follows the law of $(\ind_{\{U_1\leqslant p\}},\ind_{\{U_2\leqslant q\}})$ where $p=\mathbb{P}(\delta=1)$, $q=\mathbb{P}(\xi=1)$ and $(U_1,U_2)$ is a random vector with distribution function $D$. Unlike the case of continuous margins, the copula is not unique outside the range of the discontinuous margin distributions. In this work, we extend the copula definition conditionally on the random couple $(T,X)$ which leads to the following assumption.\\
 
 \noindent
 {\bf Assumption $(\mathcal{C})$.}\textit{ There exist a random couple $(U_1,U_2)$ with known bivariate distribution function $D:\mathbb{R}^2\to[0,1]$ such that one can find two real functions $h_0$ and $\ell_0$ taking values in $\mathbb{R}_+\times\mathbb{R}^p$ with $(\delta,\xi)$ following the same law as $(\ind_{\{U_1\leqslant h_0(T,X)\}},\ind_{\{U_2\leqslant \ell_0(T,X)\}})$ conditionally on $(T,X)$.}\\
 
 \noindent
 Note that for the sake of simplicity, we consider the distribution $D$ independent from the distribution of $(T,X)$ and leave this feature for further developments. Although the previous assumption is equivalent to the common copula definition up to some functional transformations, it facilitates the analogy with other missingness models. Among others, we can cite the common logit model with linear functions or the Heckman's model \cite{galimard2018heckman} where $(U_1,U_2)$ is a random standard normal couple.\\ 
 
 In view of the result in Proposition \ref{prop:plugin}, we shall develop an estimator $p_n$ that consistently estimates the function $h$, at least in the sense of (\ref{assumption_pn}). Since not all the censoring indicators are available, our approach is twofold. First, we solely estimate the function $\ell_0$ by fitting a maximum likelihood criterion to the distribution of $\xi$. Second, we similarly estimate $h_0$ but based on the bivariate distribution $D$ and the latter estimator of $\ell_0$. More specifically, let $\mathcal{H}$ and $\mathcal{L}$ be two classes of measurable functions to whom $h_0$ and $\ell_0$ respectively belong. Then, we define 
\begin{eqnarray}
\label{eq:likelihood1}
\widehat{\ell}_n &:=& \underset{\ell\in\mathcal{L}}{\text{argmax }}\mathbb{M}^{(2)}_n(\ell)\\ 
\nonumber&:=&  \underset{\ell\in\mathcal{L}}{\text{argmax }}\frac{1}{n}\sum_{i=1}^n\xi_i\log\left[D_2(\ell(T_i,X_i))\right] +(1-\xi_i)\log\left[1-D_2(\ell(T_i,X_i))\right]
\end{eqnarray}
and
\begin{eqnarray}
\label{eq:likelihood2}
\widehat{h}_n &:=& \underset{h\in\mathcal{H}}{\text{argmax }}\mathbb{M}^{(1)}_n(h,\widehat{\ell}_n)\\ 
\nonumber&:=& \underset{h\in\mathcal{H}}{\text{argmax}} \frac{1}{n}\sum_{\{\xi_i=1\}}\delta_i\log\left[D(h(T_i,X_i),\widehat{\ell}_n(T_i,X_i))\right] \\ 
\nonumber &&+(1-\delta_i)\log\left[D_2(\widehat{\ell}_n(T_i,X_i))-D(h(T_i,X_i),\widehat{\ell}_n(T_i,X_i))\right]
\end{eqnarray}
where $D_i$ is the $i$-th component margin distribution function.\\ 

To obtain the convergence for the estimators $\widehat{\ell}_n$ and $\widehat{h}_n$, we have to restrict the sets of candidates to not be "too large". In order to define such constrain, we introduce some notations borrowed from the theory of the weak convergence of empirical processes. First, we say that any class of measurable functions $\mathcal{F}$ is \textit{VC}(-subgraph) if the collection of all subgraphs of the functions in $\mathcal{F}$ forms a \textit{VC}-class of sets, i.e. there exists a set of $n$ points such that the set of subgraphs cannot pick out the $2^n$ subsets of the points collection \cite{Vaart1996}. Next, for any probability measure $Q$ and $\epsilon>0$, we define the covering number $N(\mathcal{F},L_2(Q),\epsilon)$ as the minimal number of $L_2(Q)-$balls of radius $\epsilon$ needed to cover $\mathcal{F}$. We say that the class $\mathcal{F}$ satisfies a uniform entropy condition if one can find $A>0$ and $\nu>0$ such that for any probability measure $Q$ and $\epsilon>0$,
\begin{eqnarray*}
N(\mathcal{F},L_2(Q),\epsilon\Vert F\Vert_{Q,2})\leq\left(\dfrac{A}{\epsilon}\right)^{2(\nu-1)},
\end{eqnarray*} 
where $0<\Vert F\Vert^2_{Q,2}=\int F^2 dQ<\infty$ and $F$ is an envelope function of the class ${\mathcal F}$. Note that any $\textit{VC}$ class of functions satisfies the previous entropy condition. Second, we define the uniform entropy integral as
\begin{eqnarray*}
J(\delta,\mathcal{F},L_2) =\int_0^\delta\sqrt{\log\sup_{\mathcal{Q}}N(\mathcal{F},L_2(Q),t\Vert F\Vert_{Q,2})}\, dt,
\end{eqnarray*} 
where $\mathcal{Q}$ is the set of all probability measures $Q$. At last, we define the expectation under $Q$ as $Qf=\int fdQ$ for any real-valued measurable function $f$. These definitions, which might seem complicated at first sight, allow proving the main proposition of this section which gives sufficient conditions so that the minimizing estimators have the correct speed of convergence.
\begin{theorem}
\label{theorem:vc}
Let us consider that $\mathcal{H}$ and $\mathcal{L}$ are two \textit{VC} classes of functions uniformly bounded in\ the finite intervals $[r_i,R_i]\subset\text{int}\left(U_i(\Omega)\right)$ respectively. We also assume that on the same intervals, $\exists m,M>0$ such that $D$ is differentiable with $\frac{\partial}{u_i}D$ uniformly bounded in $[r,R]$ for $i=1,2$ and the derivative function $u_1\to\frac{\partial}{u_2} D(u_1,u_2)$ is Lipschitz uniformly in $u_2\in(r_2,R_2)$. Then, for any non-random sequences $q_{1,n}$ and $q_{2,n}$ such that $q_{i,n}^2\psi^{(i)}(1/q_{i,n})\leq\sqrt{n}$ for $i=1,2$ and $n$ large enough with
\begin{eqnarray*}
\psi_n^{(1)}(\varepsilon) = \left(\psi_n^{(2)}(\varepsilon)+\varepsilon^2\sqrt{n}\right)^{1/2} \quad\text{and}\quad \psi_n^{(2)}(\varepsilon) =(\varepsilon+n^{-1/2})|\log(\varepsilon)|
\end{eqnarray*}
we have
\begin{eqnarray}
\label{proof:speed}
\mathbb{E}\left[(\widehat{h}_n-h_0)^2(T,X)\right]^{1/2}=O\left(\frac{1}{q_{n,1}}\right) \hspace{.2cm}\text{and}\hspace{.2cm}\mathbb{E}\left[(\widehat{\ell}_n-\ell_0)^2(T,X)\right]^{1/2}=O\left(\frac{1}{q_{n,2}}\right).
\end{eqnarray}
\end{theorem}
To illustrate the practical applicability of such a result, we can show that there exist some rates of convergence such that the estimator $\widehat{h}_n$ fits (\ref{assumption_pn}) and thus provides the correct estimator for $\widecheck{F}_n$. Indeed, one can define $q_{n,1}=O(n^\alpha)$ and $b=O(n^{-\beta})$ with $0<\alpha<\frac{1}{12}$ and $\frac{5}{6p}<\beta<\frac{1}{p}$ and check that it fits the assumptions from the Proposition \ref{prop:plugin} and Theorem \ref{theorem:vc}. It is finally worth mentioning that the \textit{VC}-class assumption is available for large classes of functions, for instance with the neural networks \cite{Karpinski1997} which we will use in the next sections for the MNAR applications.

\section{Experiments on synthetic data \label{sec:expe}}

We now report the results of a simulation study with the performances of different versions of the  Beran estimator in the MAR and MNAR settings compared with reference benchmarks computed before data deletion. This section first describes the synthetic data-generating process, followed by a description of  the analysis method  and the results.

\subsection{Data-generating Process}

The data generation is done in two steps. We first consider a standard censorship model and generate the survival data and then generate the missing censorship indicators accordingly to a given dependence structure. We here consider Heckman's model to drive the dependency between the censorship indicator and the indicator of missing data.
The covariate $X:=(X^{(1)},X^{(2)},X^{(3)})$ is a three dimensional variable uniformly distributed on $[0,1]^3$. The Weibull and Frechet distributions are taken to generate the survival time $Y$ from $X$ with the following conditional cumulative distribution functions
\begin{itemize}
    \item Weibull $F(y|X=X_i)=1-e^{-y^{a(X_i)}}$,
    \item Frechet $F(y|X=X_i)=e^{-y^{-a(X_i)}}$
\end{itemize}
where the function $a$ is linear and define by $a(X):= a_0 + a_1 X^{(1)} + a_2 X^{(2)} + a_3 X^{(3)}$. The censoring variable is independent of  $Y$ conditionally to $X$ and built with an exponential model given by $C:=\exp(b(X))$ where $b(X):= b_0 +  b_1X^{(1)}  + b_2X^{(2)} + b_3X^{(3)}$.
The non-random sampling of the missingness mechanism is hence modeled by
\begin{equation*}
    \mathbb{P}(\xi_i=1|X=X_i,T=T_i) = \phi(\ell_0(T_i,X_i)),
\end{equation*}
where $\phi$ is the standard normal cumulative distribution and $\ell_0$ is a linear function given by:
\begin{equation}
\ell_0(T_i,X_i)  = c_0 + c_1 X^{(1)} + c_2 X^{(2)} + c_3 X^{(3)} + c_4 T_i.
\label{eq:l_equation}
\end{equation}
According to Heckman's dependency model and given the values of the censorship indicator $\delta_i$, we aim to generate the values $\xi_i$ such as:
\begin{equation}
\mathbb{P}(\xi_i=1|\delta_i=1) = \frac{\phi_2(h_0(T_i,X_i), \ell_0(T_i,X_i),\rho)}{\phi(h_0(T_i,X_i))},
\label{eq:condDelta1}
\end{equation}
and 
\begin{equation}
\mathbb{P}(\xi_i=1|\delta_i=0) = \frac{\phi_2(-h_0(T_i,X_i), \ell_0(T_i,X_i),-\rho)}{1-\phi(h_0(T_i,X_i))},
\label{eq:condDelta0}
\end{equation}
where $\phi_2$  corresponds to the binormal cumulative density function, $\rho$ is the correlation coefficient and $h_0$ must satisfy 
\begin{equation*}
\mathbb{P}(\delta_i=1|T=T_i,X=X_i) = \phi(h_0(T_i,X_i)).
\end{equation*}
Therefore we derive
\begin{equation*}
h_0(T_i,X_i) = \phi^{-1}(\mathbb{P}(\delta_i=1|T=T_i,X=X_i)),
\end{equation*}
and the censoring conditional probability can be written as
\begin{eqnarray*}
\mathbb{P}(\delta_i=1|T=T_i,X=X_i)
=\frac{f(T_i|X_i) (1 - G(T_i|X_i))}{f(T_i|X_i)(1 - G(T_i|X_i)) + g(T_i|X_i)( 1 - F(T_i|X_i))}.
\label{eq:proba_delta}
\end{eqnarray*}
where $f$ and $g$ are the conditional probability density function of the survival and censoring variables. Thus, when the Weibull and the Frechet distribution are taken for the survival time, we can respectively compute  
\begin{equation*}
h_0(T_i,X_i) = \phi^{-1}\left(  \frac{a(X_i) T_i^{a(X_i) - 1}}{a(X_i) T_i^{a(X_i) - 1} + b(X_i)}\right)
\end{equation*} and
\begin{equation*}
h_0(T_i,X_i) = \phi^{-1}\left(  \frac{a(X_i)T_i^{-1-a(X_i)}}{a(X_i)T_i^{-1-a(X_i)} + b(X_i)(\exp(T_i^{-a(X_i)})-1)}\right).
\end{equation*}
 Using this function $h_0$, the expression of $\ell_0$ (\ref{eq:l_equation}) and the censoring indicators $\delta_i$'s, the values $\xi_i$ are sampled according to the conditional probability given by equations (\ref{eq:condDelta1}) and (\ref{eq:condDelta0}).\\

\noindent
\textbf{Parameter choices} : For each of the two distributions,  nine different scenarios with different level of censoring and different level of missing delta are considered. The different scenarios are summarized in Table \ref{table:different_scenarios}.

\begin{table}[H]
\centering
\begin{tabular}{lcc}
\toprule
     &  \% of censoring  & \% of missing delta  \\
 \toprule
 Scenario 1  & 25\%  & 25\% \\
 Scenario 2  & 25\%  & 50\% \\
 Scenario 3  & 25\%  & 75\% \\
 Scenario 4  & 50\%  & 25\% \\
 Scenario 5  & 50\%  & 50\% \\
 Scenario 6  & 50\%  & 75\% \\
 Scenario 7  & 75\%  & 25\% \\
 Scenario 8  & 75\%  & 50\% \\
 Scenario 9  & 75\%  & 75\% \\
\bottomrule
\end{tabular}
 \caption{Different levels of censoring and missing delta}
\label{table:different_scenarios}
\end{table}

The sets of parameters $a_j$, $b_j$, and $c_j$, chosen to generate these nine different scenarios for both distributions, are described in the supplementary material. For all the scenarios, we use $\rho$ fixed at 0, 0.25, 0.5, and 0.75 to respectively simulate MAR, light, medium, and heavy MNAR settings from a bivariate normal distribution according to the Heckman’s dependency model between $\delta$ and $\xi$. Our simulations are based on \textit{i.i.d.} samples $\mathcal{D}^k = \{(T^k_i,X^k_i,\delta^k_i,\xi^k_i)\}_{i=1}^n$ of size $n = 2000$ for $k = 1, \dots,N$ sample iterations with $N =200$. All the experiments can be replicated using  the source code publicly available at the URL \url{https://github.com/GoudetOlivier/Survival_estimation_MNAR}. 

\subsection{Methods Analysis}
We hereafter compare the overall performances of the Beran estimator associated with different estimation and imputation methods for the missing censoring indicators.

\subsubsection*{Without Data Deletion} 
For comparative reasons with the best-expected performances, we consider data without the missingness mechanism. To do so, we compute: 
\begin{itemize}
    \item an oracle version where $p_n(T_i,X_i)$ is equal to the \textit{true} conditional probability $p(T_i,X_i) = \mathbb{P}(\delta_i=1|T=T_i,X=X_i)$.
    \item the standard Beran estimator.
\end{itemize}

\subsubsection*{Naive Approach} 
We compute the standard Beran estimator considering only fully observed data by removing the individuals with unobserved censoring indicators.

\subsubsection*{MAR} 
We compute two estimators learned in the MAR setting : 
\begin{itemize}
    \item the first one is the kernel  estimator proposed by S. Subramanian \cite{Subramanian2006}, where $p_n(T_i,X_i)$ is given by  $\frac{\xi_i\delta_i}{\hat{\pi}(T_i,X_i)}$. Here $\hat{\pi}(T_i,X_i)$ is a kernel estimate of $\mathbb{P}(\xi_i=1|T=T_i,X=X_i)$ computed as 
\begin{equation}
    \hat{\pi}(T_i,X_i) = \frac{\sum_{i=1}^nK_{\bar{b}}(x-X_i, t-T_i) \xi_i}{\sum_{i=1}^nK_{\bar{b}}(x-X_i, t-T_i)}.
    \label{eq:subramanian}
\end{equation}

The choice of the kernel function $K_{\bar{h}}$ and the bandwidth $\bar{b}$ are described in section \ref{sec:weight_estimation} below. 

\item the second one is a neural network estimator where $p_n(T_i,X_i)$ is equal to $\phi(\hat{h}_{\mathbf{\alpha}}(T_i,X_i))$, with  $\hat{h}_{\alpha}$ a regression neural network and a set of parameters $\alpha$. It is composed of three hidden layers with 200, 200 and 100 neurons. The set of parameters $\alpha$ is trained by gradient descent for $n_{\textit{iter}} = 1000$ epochs with a batch size of 100 data points using Adam optimizer \cite{kingma2014adam}, in order to maximize the following log-likelihood of the probit model on the observed data:
\begin{eqnarray*}
    \mathcal{L}^{MAR}(\mathbf{\alpha}) &=& \sum_{\{i;\xi_i=1,\delta_i=1 \}} \text{log} \  \phi(\hat{h}_{\mathbf{\alpha}}(T_i,X_i)) + \sum_{\{i;\xi_i=1,\delta_i=0 \}} \text{log} (1 -   \phi(\hat{h}_{\mathbf{\alpha}}(T_i,X_i))).
\end{eqnarray*}
Therefore $\phi(\hat{h}_{\mathbf{\alpha}}(T_i,X_i))$ provides 
an estimated value of the conditional probability $\mathbb{P}(\delta_i=1|T=T_i,X=X_i)$  learned in the MAR setting. 

\end{itemize}

\subsubsection*{MNAR} 
We compute two versions of  MNAR Beran estimators with neural networks :
\begin{itemize}
    \item in the first version, $p_n(T_i,X_i)$ is always equal  $\phi(\hat{h}_{\mathbf{\alpha}}(T_i,X_i))$, an estimated value of the conditional probability $\mathbb{P}(\delta_i=1|T=T_i,X=X_i)$ learned in the MNAR setting.
    \item in the second version, $p_n(T_i,X_i)$ 
 is equal to $\delta_i$ when $\delta_i$ is observed and equal to $\phi(\hat{h}_{\mathbf{\alpha}}(T_i,X_i))$ otherwise. 
\end{itemize}
Following the estimating procedure described in (\ref{eq:likelihood1}) and (\ref{eq:likelihood2}), two probit models $\phi(\hat{\ell}_{\mathbf{\beta}}(T_i,X_i))$ and $\phi(\hat{h}_{\mathbf{\alpha}}(T_i,X_i))$  are learned, where $\hat{\ell}_{\mathbf{\beta}}$  and $\hat{h}_{\mathbf{\alpha}}$ are two regression neural networks. Each of them is composed of three hidden layers with 200, 200 and 100 neurons. These two neural networks are respectively parametrized by the sets of parameters $\mathbf{\alpha}$ and $\mathbf{\beta}$, which are  learned by gradient descent in two steps for $n_{\textit{iter}} = 1000$ epochs with a batch size of 100 data points using Adam optimizer \cite{kingma2014adam}.
In the first step, we aim to find the set of parameters $\beta$ maximizing the log likelihood of the distribution of $\xi$:
 \begin{equation}
     \mathcal{L}^{\xi}(\beta) = \sum_{\{i;\xi_i=1 \}} \text{log} \  \phi(\hat{l}_{\mathbf{\beta}}(T_i,X_i)) + \sum_{\{i;\xi_i=0 \}} \text{log} (1 -   \phi(\hat{l}_{\mathbf{\beta}}(T_i,X_i))).
 \end{equation}
 Once the function $\hat{l}_{\mathbf{\beta}}$ is learned, we use it to find $\mathbf{\alpha}$  maximizing the following log-likelihood of the joint bivariate probit model \cite{van1981demand,marra2013penalized,galimard2018heckman}:
 \begin{align}\label{eq::alpha_minim}
     \mathcal{L}^{MNAR}(\mathbf{\alpha}) &= \sum_{\{i;\xi_i=1,\delta_i=1 \}} \text{log} \  \phi_2(\hat{h}_{\mathbf{\alpha}}(T_i,X_i),\hat{l}_{\mathbf{\beta}}(T_i,X_i),\rho)
    \\ \nonumber & \ \  \ \ + \sum_{\{i;\xi_i=1,\delta_i=0 \}} \text{log} \  \phi_2(-\hat{h}_{\mathbf{\alpha}}(T_i,X_i),\hat{l}_{\mathbf{\beta}}(T_i,X_i),-\rho)
 \end{align}
The initial learning rate is set to 0.001 for $\mathbf{\alpha}$ and $\mathbf{\beta}$. The gradient  of $\phi_2$ with respect to $\mathbf{\alpha}$ is computed with the \textit{torch-mvnorm} package\footnote{\url{https://github.com/SebastienMarmin/torch-mvnorm}}
and using the formula given by \cite{marmin2015differentiating}. 

\subsection{Bandwidth Selection \label{sec:weight_estimation}}
Once the set $\{p_n(T_i,X_i),\,i=1,\ldots,n\}$ is computed, it remains to compute the Beran weights for the various estimators (cf. equation \ref{eq:GBE_estimator}). We here use the same bi-quadratic kernel function $K(x) = (15/16)(1-x^2)^2 \ind_{\{|x| \leq 1\}}$, to compute the weights and the estimated probability $\hat{\pi}(T_i,X_i)$ for the Subramanian type estimator (cf. equation \ref{eq:subramanian}). For each method and each sample, the bandwidth value $b_{\text{best}}$ is selected in the grid $\{0.1, 0.125, 0.15, \dots, 0.3\}$  that minimizes the following leave-one-out cross-validated criterion:
\begin{equation*}
    b_{\text{best}} = \underset{h}{\text{argmin}} \left[ \sum_{i=1}^{n} \sum_{j=1}^{n} \Delta_{ij} \left( I(T_i \leq T_j) -  F^{(-i)}_{h,n} (T_j|X_i) \right)^2 \right],
    \label{eq:kernel}
\end{equation*}
where $F^{(-i)}_{h,n} (\cdot|X_i)$ is the generalized Beran estimator computed with bandwidth $h$ and using the sample $\{(T_j,X_j,p_n(T_j,X_j)),\,j=1,\ldots,n,\,j\neq i\}$. The weights $\Delta_{ij}$ are equaled to  $p_n(X_i,T_i)$ if $T_i \leq T_j$ or $\Delta_{ij} = p_n(X_j,T_j)$ if $T_j < T_i$. Each weight $\Delta_{ij}$ corresponds to the probability that the value of the indicator $\ind_{\{T_i \leq T_j\}}$ gives an unambiguous correct value for the indicator $\ind_{\{Y_i \leq Y_j\}}$, which contains the corresponding true event times $(Y_i,Y_j)$. When $p_n(X_j,T_j) = \delta_i$, this criterion is equivalent to that proposed by \cite{geerdens2018conditional}.

\subsection{Performance Evaluation \label{sec:perf_eval}}

In each data-generating scenario, the performance of
each method is assessed by comparing the different estimators with the \textit{true} conditional distribution function of the survival times given by $F(t|X=x)=1-e^{-t^{a(x)}}$ for the Weibull distribution and $F(t|X=x)=e^{-t^{-a(x)}}$ for the Frechet distribution. The comparisons are done for $27$ different values $x_{test}\in\{0.25,0.5,0.75\}^3$  of the covariate vector $X$ in $[0,1]^3$. For each data sample $\mathcal{D}^k$, a global mean integrated squared error (MISE) score  between $F$ and $\widecheck{F}^k_n$ is computed on $M=100$ equidistant design points $t_m$ between $\displaystyle\min_{i=1,...,n}\{Y_i\}$ and $\displaystyle\max_{i=1,...,n}\{Y_i\}$ such that
\begin{eqnarray*}
 \text{MISE}(\mathcal{D}^k) = \frac{1}{27M}\sum_{t_m,\,x_{test}}  \left[F(t_m|X=x_{test}) - \widecheck{F}^{k}(t_m|X=x_{test})\right]^2,
\label{eq:mise_global}
\end{eqnarray*}
where $\widecheck{F}_n^{k}$ is the conditional distribution function estimator computed with the best bandwidth value for the sample $\mathcal{D}^k$.

\subsection{Results}

\begin{sidewaysfigure}
\centering
    \includegraphics[scale=0.2]{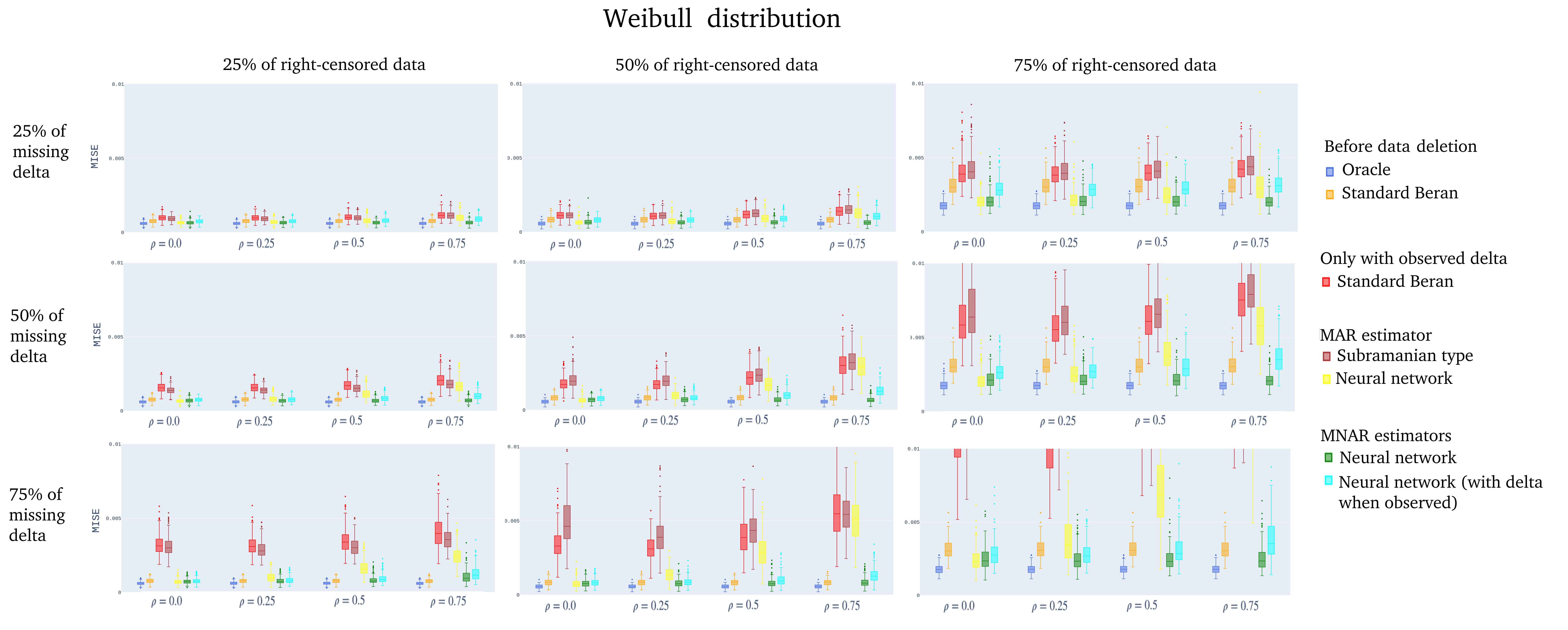}
    \centering
    \caption{Different Beran estimators built from synthetic data generated with the Weibull distribution}
    \label{fig:weibull}
\end{sidewaysfigure}

\begin{sidewaysfigure}
\centering
    \includegraphics[scale=0.2]{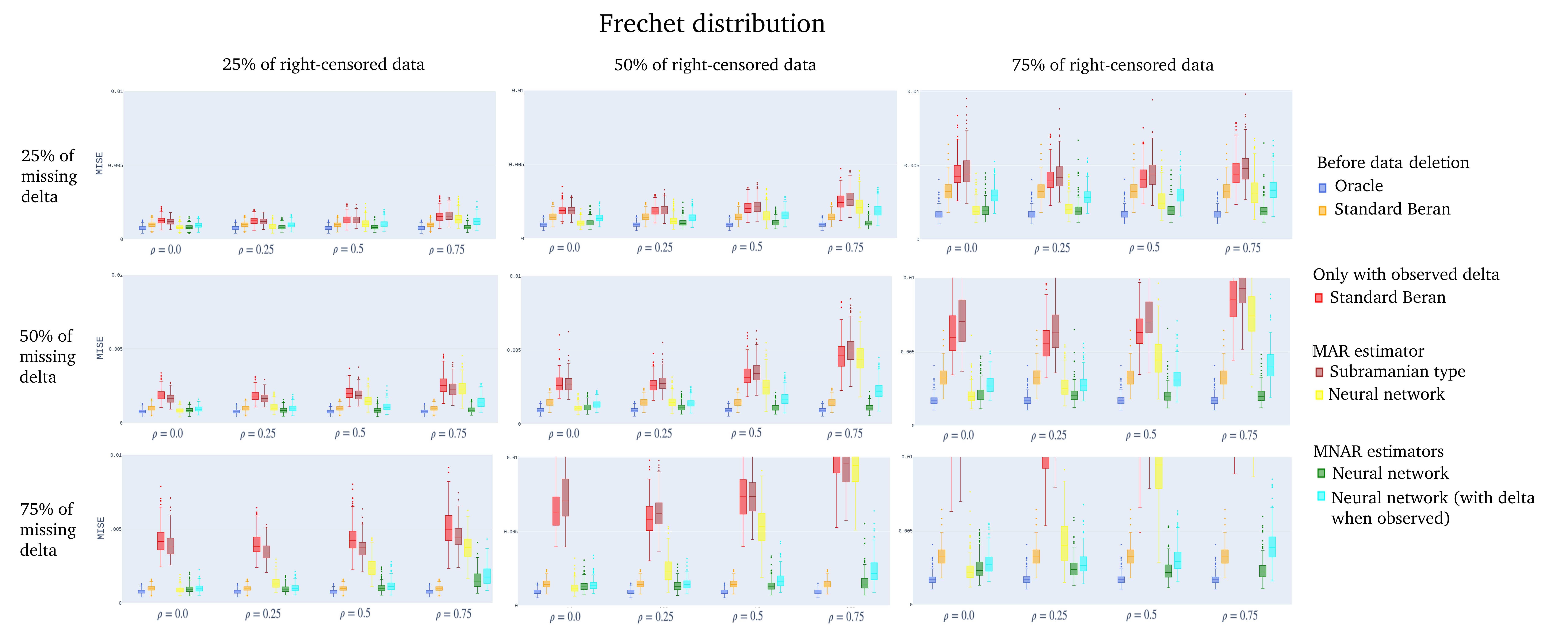}
    \centering
    \caption{Different Beran estimators built from synthetic data generated with the Frechet distribution}
    \label{fig:frechet}
\end{sidewaysfigure}

Figures \ref{fig:weibull} and \ref{fig:frechet} show boxplots of the simulation study results with data generated according to the Weibull and Frechet distributions with different levels of right-censored data, missing delta, and dependency between $\delta$ and $\xi$, respectively. Each boxplot displays the MISE score given by equation (\ref{eq:mise_global}) computed for $N=200$ independent replication samples.

\noindent
First, the oracle version (in blue) unsurprisingly performs best for all scenarios and all values of $\rho$. In particular, it is better than the standard Beran estimator (in orange), where each sample $p_n(T_i,X_i)$ is equal to $\delta_i$, which experimentally confirms the theorem \ref{theorem_nesti} showing that the variance is minimal for $p_n(T_i,X_i)=\mathbb{P}(Y_i \leq C_i|T_i,X_i)$.\\ 
Second, the standard Beran estimator built before data deletion (in orange), with the entire data sample, is always better than the same standard Beran estimator built only with the fraction of the sample corresponding to the observed delta (in red). As expected, these differences become more significant as the percentage of missing delta increases.\\
Third, comparing the two MAR estimators, the Subramanian type (brown) and the neural network (yellow), we observe that the neural network always performs better, suggesting that the neural network can better capture the dependence between survival times and covariates, than the kernel model.\\  
Fourth, we confirm that the MAR and MNAR neural network estimators (in yellow and green, respectively) obtain the same results when $\rho = 0.0$ (in the MAR framework). However, as $\rho$ increases, the results of the MAR neural network estimator deteriorate rapidly while the MNAR neural network estimator remains very robust for all scenarios. The difference between these two estimators becomes increasingly important as $\rho$ increases, and more so as the percentage of right-censored data and the missing delta are large.
Lastly, when comparing the two versions of the neural networks estimator built in the MNAR setting, we observe that the version in green, where $p_n(T_i,X_i)$ is always equal  to $\phi(\hat{h}_{\mathbf{\alpha}}(T_i,X_i))$, the estimated value of the conditional probability $\mathbb{P}(\delta_i=1|T=T_i,X=X_i)$, is always better than the second version in cyan blue, where $p_n(T_i,X_i)$ 
 is equal to $\delta_i$ when $\delta_i$ is observed, and equal to $\phi(\hat{h}_{\mathbf{\alpha}}(T_i,X_i))$ when $\delta_i$ is not observed. It highlights for these experiments, that the estimated values of the conditional probabilities $\mathbb{P}(\delta_i=1|T=T_i,X=X_i)$ with the neural network are sufficiently precise so that it is not interesting to replace them with the censoring indicators when they are observed.

\section{Experiments on real data \label{sec:expe_real}}
We apply to a real dataset the estimators associated with different imputation methods for the missing censoring indicators. The dataset is provided by the National Cancer Institute (NIH) in the United States and gathers mortality data from 2015 for the prostate, lung, colorectal and ovarian cancer. The death certificates are the primary source of information for the causes of death. A final review by an independent Death Review Committee determines the cause of death for patients diagnosed with cancer, sometimes resulting in a different conclusion if the death certificate was ambiguous. Thus, occasionally a death is not attributed to a cancer and the censoring indicator is left missing. Here, we selected a panel of 8294 participants from ten different study centers among those who had confirmed primary invasive prostate cancer and where no missing data were observed in the covariates presented below. The variables we consider for each patient are :
\begin{itemize}
\item $T$ the number of days  from trial entry  until mortality exit date. This is the day of death or the day last known alive. 
\item $X:=(X^{(1)},X^{(2)},X^{(3)})$ a vector of continuous covariates, with $X^{(1)}$ the patient age in years, $X^{(2)}$ the number of packs smoked per day times the number of years of smoking and $X^{(3)}$ the body mass index (BMI) of the patient. 
\item $\delta$ the occurence of death (1-dead, 0-patient is known alive at the end of the study). 
\item $\xi$ the indicator of missing censoring indicator $\delta$  (1-confirmation of patient status dead or alive, 0-not confirmed).
\end{itemize}
In order to compute the Beran estimators, all the covariates and survival times are normalized in order to take values in $[0,1]$. We computed the following four different estimators 



\begin{figure}[H]
\centering
    \includegraphics[scale=0.4]{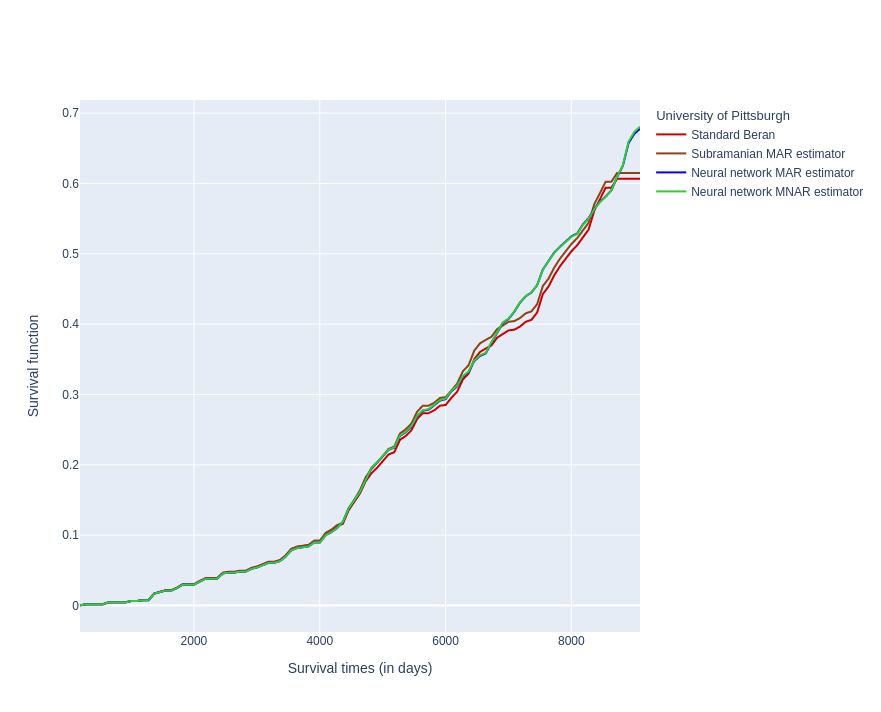}
     \caption{Survival function estimators built with the University of Pittsburgh center data (950 participants, 51\% of right-censored data and 3\% of missing censoring indicator). }
     \label{fig:center_8}
\end{figure}

\begin{itemize}
\item the standard Beran estimator considering only fully observed data by removing patients with unobserved censoring indicator (when $\xi = 0)$.
\item the kernel estimator proposed by S. Subramanian \cite{Subramanian2006} learned in the MAR setting. 
\item  the neural network estimator  learned in the MAR setting described in the last section.
\item the neural network estimator learned in the MNAR setting as presented in the experiment section. Since the covariance parameter $\rho$ is unknown, we adapt the approach in (\ref{eq::alpha_minim}) by also maximizing over the values of $\rho\in(-1,1)$.
\end{itemize}
For each method, we keep  the same set of parameters and the same bandwidth selection procedure as detailed in the last section and each survival function estimation is averaged over 200 replications (because of stochastic effects related to the random initialization of the neural network weights). When there is almost no missing censoring indicators and enough observed data, the four different survival estimators are very close to each other. As an example, in Figure \ref{fig:center_8} are displayed four different estimators based on data from the center of the University of Pittsburgh (950 participants, 51\% of right-censored data and 3\% of missing censoring indicator) for $x_{test} = \{0.5,0.5,0.5\}$. This validates to some extent the consistency of the different estimators in the favorable scenario with few missingness. 

\begin{figure}[H]
\centering
    \includegraphics[scale=0.4]{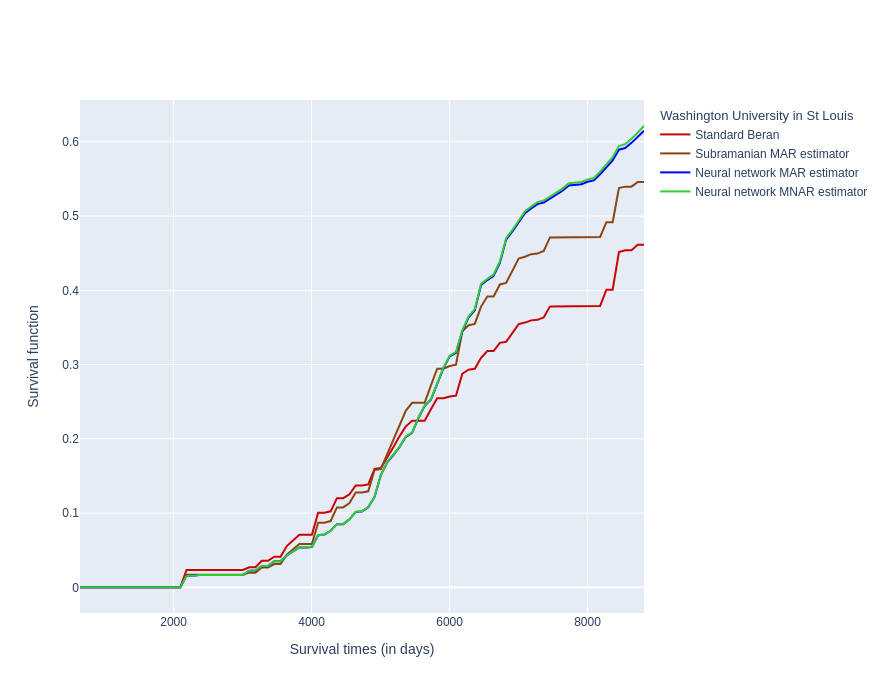}
     \caption{Survival function estimators built with the Washington University center data in St. Louis (696 participants, 65\% of right-censored data  and 24\% of missing censoring indicator). }
     \label{fig:center_6}
\end{figure}

\noindent
However, when there are more right-censored data and more missing censoring indicators, we observe for some study centers an higher discrepancy between the different approaches. In Figure \ref{fig:center_6} are displayed the four estimators based on the data from the Washington University center in St. Louis (696 participants, 65\% of right-censored data  and 24\% of missing censoring indicator) for $x_{test} = \{0.5,0.5,0.5\}$.

\noindent
For high values of the survival times, when there are fewer surviving patients at these time step, the neural network estimators (green and blue lines) lead to a smoother estimate of the survival function due to the use of the learned partial censorship indicator $P$ instead of the binary disease free survival indicator $\delta$. The two neural network estimators learned in the MAR (blue line) and MNAR setting (green  line) are also very close for these data. It can be explained by the fact that the averaged estimate of the covariance parameter $\rho$ is below 0.1 in absolute value when the neural network is trained in the MNAR setting on these data. This might suggest that the missingness pattern is actually independent for those data but in the same time, emphasizes the usefulness of the MNAR approach when no a-priori information is available on the dependence structure.

\section{Discussion and Perspectives \label{sec:conclusion}}

The main contribution of this paper is to propose a consistent approach to construct efficient estimators through the idea of generalized censoring indicators. In the context of missing not at random censoring indicators, a plug-in estimator is built, leveraging machine learning techniques to learn a complex model of the probability of censorship from individual's covariates. Besides the theoretical analysis of the approach, the proposed methods are experimentally validated on synthetic and real data. Works are currently in progress to extend the maximization step to unknown copula function. The results on the real data have indeed shown some appealing behaviors in this direction and we expect to obtain the theoretical guarantees in the future work.



%

\bibliography{bibli.bib} 
\bibliographystyle{plain}

\newpage
\appendix

\section{Supplementary material for : "Survival Estimation for Missing not at Random Censoring Indicators based on Copula Models"}

This appendix regroups the proofs of the main article and also provides the parameters used in the various scenarios from the experiments for the synthetic data generation according to the censoring and missing rates.

\textbf{Proof of Lemma \ref{lambda}}.\\ 
The proof here is similar to that of Lemma 3.6 in \cite{Dikta1998}. 
According to the mean value theorem, we have for any $t\in[\tau_0,\tau_1]$
\begin{eqnarray}
\label{proof_lambda}
\nonumber\left|1-\widehat{F}_n(t|x)-\exp(-\widehat{\Lambda}_n(t|x))\right|&\leq& \left|r(t)\left[-\log(1-\widehat{F}_n(t|x))-\widehat{\Lambda}_n(t|x)\right]\right|\\
&\leq&\left|-\log(1-\widehat{F}_n(t|x))-\widehat{\Lambda}_n(t|x)\right|
\end{eqnarray}
where $r(t)$ lies between $1-\widehat{F}_n(t|x)$ and $\exp(-\widehat{\Lambda}_n(t|x))$. For any $0<y<1$, we have the inequality $0<-\log(1-y)-y<\dfrac{y^2}{1-y}$. This yields in (\ref{proof_lambda}) to
\begin{eqnarray*}
&&\hspace{-2cm}\left|1-\widehat{F}_n(t|x)-\exp(-\widehat{\Lambda}_n(t|x))\right|\\
&\leq&\sum_{T_i\leq t}P_iW_b(x-X_i)^2\left[\dfrac{1}{(1-H_n(T_i|x))^2}+\dfrac{1}{(1-H_n(T_i^-|x))(1-H_n(T_i|x))}\right]\\
&\leq&2\dfrac{\sum_{T_i\leq t}P_iW_b(x-X_i)^2}{(1-H_n(\tau_1|x))^2}\\
&\leq&2\Vert K\Vert_\infty\dfrac{\sum_{T_i\leq t}P_iW_b(x-X_i)}{nh^pf_n(x)(1-H_n(\tau_1|x))^2}
\end{eqnarray*}
and the lemma follows.\hfill$\Box$\\

\bigskip
\noindent
\textbf{Proof of the Proposition \ref{prop:asrep}}.\\ 
From the proof of Theorem 3.1 in \cite{Escobar2019}, it follows that
\begin{eqnarray*}
\Lambda_n(t|x)-\Lambda(t|x)=\sum_{i=1}^nW_b(x-X_i)\ell(t,T_i,\delta_i|x)+\mathcal{O}_\mathbb{P}((nb^p)^{-3/4} |\log b|^{3/4})
\end{eqnarray*}
where $\ell$ is given by
\begin{eqnarray*}
\ell(t,T_i,\delta_i|x)&=&\int_{-\infty}^t\dfrac{\ind_{\{T_i < s\}}-H(s|x)}{(1-H(s^-|x))^2}dH^u(s|x)+\dfrac{\ind_{\{T_i \leq t,\delta_i=1\}}-H^u(t|x)}{1-H(t|x)}\\
&&-\int_{-\infty}^t\dfrac{\ind_{\{T_i \leq s,\delta_i=1\}}-H^u(s|x)}{(1-H(s|x))^2}dH(s|x)\\
&=&\widehat{\ell}(t,T_i,\delta_i|x)+\int_{-\infty}^t\dfrac{\ind_{\{T_i < s\}}-H(s|x)}{(1-H(s|x))^2}d(H^u-H_p^u)(s|x)-\dfrac{(P_i-\delta_i)\ind_{\{T_i\leq t\}}}{1-H(T_i|x)}.
\end{eqnarray*}
It thus only remain to derive the almost sure representation for $\widehat{\Lambda}_n-\Lambda_n$. Hence, we have by definition that
\begin{eqnarray*}
\widehat{\Lambda}_n(t|x)-\Lambda_n(t|x)&=&\int_{-\infty}^t\dfrac{d(\widehat{H}_n^u-H^u_n)(s|x)}{1-H_n(s^-|x)}\\
&=&\sum_{i=1}^nW_b(x-X_i)\dfrac{(P_i-\delta_i)\ind_{\{T_i\leq t\}}}{1-H(T_i|x)}\\
&&+\int_{-\infty}^t\dfrac{1}{1-H_n(s|x)}-\dfrac{1}{1-H(s|x)}d(\widehat{H}_n^u-H^u_n)(s|x)\\
&&+\int_{-\infty}^t\dfrac{1}{1-H_n(s^-|x)}-\dfrac{1}{1-H_n(s|x)}d(\widehat{H}_n^u-H^u_n)(s|x)\\
&=:&\sum_{i=1}^nW_b(x-X_i)\dfrac{(P_i-\delta_i)\ind_{\{T_i\leq t\}}}{1-H(T_i|x)}+R_{n,1}(t)+R_{n,2}(t).
\end{eqnarray*}
Clearly, for any $t\in\mathbb{R}$ the estimators $H_n(t)$ and $H_n(t^-)$ might only differs of one jump of size $(nb^p)^{-1}$, which gives us
\begin{eqnarray}
\label{proof_jump}
\nonumber\left|\dfrac{1}{1-H_n(s^-|x)}-\dfrac{1}{1-H_n(s|x)}\right|&=&\dfrac{\left|H_n(s|x)-H_n(s^-|x)\right|}{(1-H_n(s^-|x))(1-H_n(s|x))}\\
&\leq& \dfrac{\left|H_n(s|x)-H_n(s^-|x)\right|}{(1-H_n(\tau_1|x))^2}=\mathcal{O}_\mathbb{P}((nb^p)^{-1})
\end{eqnarray}
where the last term is uniform in $s\in\mathbb{R}$, yielding that $\sup_{t\in[\tau_0,\tau_1]}|R_{n,2}(t)|=\mathcal{O}_\mathbb{P}((nb^p)^{-1})$ for $n$ large enough, since we have almost surely $H_n(\tau_1|x)\to H(\tau_1|x)<1$ as $n\to+\infty$.\\ 

Next, we have
\begin{eqnarray*}
R_{n,1}(t)&=&\int_{-\infty}^t\dfrac{1}{1-H(s|x)}-\dfrac{1}{1-H_n(s|x)}d(H_p^u-H^u)(s|x)\\
&&+\int_{-\infty}^t\dfrac{1}{1-H(s|x)}-\dfrac{1}{1-H_n(s|x)}d(H^u-H_n^u)(s|x)\\
&&+\int_{-\infty}^t\dfrac{1}{1-H(s|x)}-\dfrac{1}{1-H_n(s|x)}d(\widehat{H}^u_p-H^u_p)(s|x)\\
&=:&\int_{-\infty}^t\dfrac{1}{1-H(s|x)}-\dfrac{1}{1-H_n(s|x)}d(H_p^u-H^u)(s|x)+R_{n,3}(t)+R_{n,4}(t).
\end{eqnarray*}
In order to uniformly bound the remaining terms $R_{n,3}(t)$ and $R_{n,4}(t)$, we again refer to the proof of Theorem 3.1 in \cite{Escobar2019}. Indeed $H_n$, $H_n^u$ and $\widehat{H}_n^u$ share the same asymptotic behavior as stated in Lemma 3.1, that is 
\begin{eqnarray*}
\label{rate1}
\sup_{t\in\mathbb{R}}|H_n(t|x)-H(t|x)|&=&\mathcal{O}_\mathbb{P}\left((nb^p)^{-1/2}|\log b|^{1/2}\right),\nonumber\\
\sup_{t\in\mathbb{R}}|H^u_n(t|x)-H^u(t|x)|&=&\mathcal{O}_\mathbb{P}\left((nb^p)^{-1/2}|\log b|^{1/2}\right),\\
\sup_{t\in\mathbb{R}}|\widehat{H}^u_n(t|x)-H^u_p(t|x)|&=&\mathcal{O}_\mathbb{P}\left((nb^p)^{-1/2}|\log b|^{1/2}\right)\nonumber
\end{eqnarray*}
where the arguments for $\widehat{H}^u_n$ are the same than that of $H_n$ and $H^u_n$. Since $H$, $H^u$ and $H^u_p$ are continuous, it yields from the proof of Lemma 2.1 in \cite{VanKeilegom1997} that
\begin{eqnarray*}
&&\sup_{t\in[\tau_0,\tau_1]}|R_{n,3}(t)|=\mathcal{O}_\mathbb{P}((nb^p)^{-3/4} |\log b|^{3/4})\\
\text{and}&&\sup_{t\in[\tau_0,\tau_1]}|R_{n,4}(t)|=\mathcal{O}_\mathbb{P}((nb^p)^{-3/4} |\log b|^{3/4}).
\end{eqnarray*}
Next, it is sufficient to see that 
\begin{eqnarray*}
&&\int_{-\infty}^t\dfrac{1}{1-H(s|x)}-\dfrac{1}{1-H_n(s|x)}d(H^u-H_p^u)(s|x)\\
&=&\int_{-\infty}^t\dfrac{H_n(s|x)-H(s|x)}{(1-H(s|x))^2}d(H^u-H_p^u)(s|x)+\mathcal{O}_\mathbb{P}\left((nb^p)^{-1}|\log h|\right)
\end{eqnarray*}
since
\begin{eqnarray*}
\dfrac{1}{1-H(s|x)}-\dfrac{1}{1-H_n(s|x)}&=&\dfrac{H_n(s|x)-H(s|x)}{(1-H(s|x))^2}-\dfrac{(H_n(s|x)-H(s|x))^2}{(1-H(s|x))^2(1-H_n(s|x))}\\
&=&\dfrac{H_n(s|x)-H(s|x)}{(1-H(s|x))^2}+\mathcal{O}_\mathbb{P}\left((nb^p)^{-1}|\log b|\right).
\end{eqnarray*}
Finally
\begin{eqnarray*}
&&\widehat{\Lambda}_n(t|x)-\Lambda_n(t|x)\\
&=&\sum_{i=1}^nW_h(x-X_i)\dfrac{(P_i-\delta_i)\ind_{\{T_i\leq t\}}}{1-H(T_i|x)}-\int_{-\infty}^t\dfrac{H_n(s|x)-H(s|x)}{(1-H(s|x))^2}d(H^u-H_p^u)(s|x)\\
&&+\mathcal{O}_\mathbb{P}((nb^p)^{-3/4} |\log b|^{3/4})
\end{eqnarray*}
and the result follows.
\hfill$\Box$\\

\noindent
\textbf{Proof of the Corollary \ref{corollary_Fn}}.\\ 
According to Lemma \ref{lambda}, we have 
\begin{eqnarray*}
&&\widehat{F}_n(t|x)-F(t|x)=\exp(-\Lambda(t|x))-\exp(-\widehat{\Lambda}_n(t|x))+\mathcal{O}_\mathbb{P}((nb^p)^{-1}).
\end{eqnarray*}
It thus only remain to use the almost sure representation for $\widehat{\Lambda}_n-\Lambda$ as shown in Proposition \ref{prop:asrep}. Indeed, by Taylor's expansion
\begin{eqnarray*}
\exp(-\Lambda(t|x))-\exp(-\widehat{\Lambda}_n(t|x))&=&(\widehat{\Lambda}_n(t|x)-\Lambda_n(t|x))\exp(-\Lambda(t|x))(1+o_\mathbb{P}(1))\\
&=&(1-F(t|x))(\widehat{\Lambda}_n(t|x)-\Lambda_n(t|x))(1+o_\mathbb{P}(1)).
\end{eqnarray*}
\hfill$\Box$\\

\noindent
\textbf{Proof of the Theorem \ref{theorem_nesti}}.\\ 
Here the proof follows the same than that of Theorem 3.2 in \cite{Escobar2019}. We thus mostly refer to this proof by adapting the milestones to our estimator. We also keep the same notations for convenience. Thanks to the assumption $(\mathcal{E})$, we have that $H^u=H^u_p$ and by integration by parts, we can rewrite the function $\widehat{g}$ given by
\begin{eqnarray}
\label{proof_g}
\widehat{g}(t,T,\delta,P|x)&=&(1-F(t|x))\widehat{\ell}(t,T,\delta,P|x)\nonumber\\
&=&\left\{\dfrac{\ind_{\{\delta=1,T\leq t\}}}{1-H(T|x)}-\int_{-\infty}^{T\nonumber\wedge t}\dfrac{dH^u(s|x)}{(1-H(s|x))^2}+\dfrac{(P-\delta)\ind_{\{T\leq t\}}}{1-H(T|x)}\right\}\\
&=&g(t,T,\delta|x)+(1-F(t|x))\dfrac{(P-\delta)\ind_{\{T\leq t\}}}{1-H(T|x)}
\end{eqnarray}
and define the sequence of classes $\widehat{\mathcal{F}}_n$ with functions taking values in $E=\mathbb{R}\times\{0,1\}\times[0,1]\times\mathcal{S}_X$ as
\begin{eqnarray*}
\widehat{\mathcal{F}}_n &=& \left\{(u,v,v',w)\rightarrow \widehat{f}_{n,t}(u,v,v',w), \, t\in[\tau_0,\tau_1]\right\} \\
&=& \left\{(u,v,v',w)\rightarrow\sqrt{h^p}K_h(x-w)\widehat{g}(t,u,v,v'|x),\, t\in[\tau_0,\tau_1]\right\},
\end{eqnarray*}
embedded with the envelope function  $E_n(u,v,v',w)=\sqrt{h^p}K_b(x-w)M$, $M>0$ being an appropriate constant since $\widehat{g}(.|x)$ is uniformly bounded.\\

The weak convergence of the stochastic process (\ref{process}) follows from the four conditions (6.6), (6.7), (6.8) and (6.9) as described in \cite{Escobar2019} page 26. However, (6.6) and (6.9) has already been proven and it only remains to show (6.7) and (6.8), that is
\begin{eqnarray}
\label{ST1}
\sup_{|t-s|\leq \delta_n} Q(\widehat{f}_{n,t}-\widehat{f}_{n,s})^2 &\longrightarrow& 0 \mbox{ for every $\delta_n \searrow 0$,}\\
\label{ST2}
J(\delta_n, \widehat{\mathcal{F}}_n, L_2)&\longrightarrow& 0 \mbox{ for every $\delta_n \searrow 0$.}
\end{eqnarray}
where $Q$ denotes the law of the vector $(T,\delta,X)$. In order to prove (\ref{ST1}), without lost of generality, we have for any $s<t\in\mathbb{R}$
\begin{eqnarray*}
(\widehat{f}_{n,t}-\widehat{f}_{n,s})(u,v,v',w)=(f_{n,t}-f_{n,s})(u,v,w)+\ind_{\{s<u\leq t\}}\dfrac{v'-v}{1-H(u|x)}
\end{eqnarray*}
that give us by convexity
\begin{eqnarray*}
[(\widehat{f}_{n,t}-\widehat{f}_{n,s})(u,v,v',w)]^2&\leq& 2[(f_{n,t}-f_{n,s})(u,v,v',w)]^2+8\dfrac{\ind_{\{s<u\leq t\}}}{(1-H(\tau_1|x))^2}.
\end{eqnarray*}
Hence, we are able to obtain the following inequality
\begin{eqnarray*}
Q(\widehat{f}_{n,t}-\widehat{f}_{n,s})^2\leq 2Q(f_{n,t}-f_{n,s})^2+8\dfrac{\mathbb{P}(s<T\leq t)}{(1-H(\tau_1|x))^2}=2P(f_{n,t}-f_{n,s})^2+8\dfrac{H(t)-H(s)}{(1-H(\tau_1|x))^2}.
\end{eqnarray*}
Since we already have that $\displaystyle\sup_{|t-s|\leq \delta_n} Q(f_{n,t}-f_{n,s})^2=o(1)$ in \cite{Escobar2019}, (\ref{ST1}) follows by the uniform continuity of the distribution function $H$. To prove (\ref{ST2}), we use the results for the class of function $\mathcal{F}_n$ introduced in \cite{Escobar2019} and apply them to $\widehat{\mathcal{F}}_n$.  Between $\widehat{\mathcal{F}}_n$ and $\mathcal{F}_n$, we have the relation
\begin{eqnarray*}
\widehat{\mathcal{F}}_n&=&\left\{(u,v,v',w)\rightarrow\sqrt{b^p}K_b(x-w)\left[g(t,u,v|x)+(1-F(t|x))\dfrac{(v-v')\ind_{\{u\leq t\}}}{1-H(u|x)}\right],\, t\in[\tau_0,\tau_1]\right\}\\
&\subset&\mathcal{F}_n+\left\{(u,v,v',w)\rightarrow\sqrt{b^p}K_b(x-w)(1-F(t|x))\dfrac{(v-v')\ind_{\{u\leq t\}}}{1-H(u|x)},\, t\in[\tau_0,\tau_1]\right\}\\
&\subset&\mathcal{F}_n+\left\{(u,v,v',w)\rightarrow\sqrt{b^p}K_b(x-w)(1-F(t|x))\dfrac{v\ind_{\{u\leq t\}}}{1-H(u|x)},\, t\in[\tau_0,\tau_1]\right\}\\
&&+\left\{(u,v,v',w)\rightarrow-\sqrt{b^p}K_b(x-w)(1-F(t|x))\dfrac{v'\ind_{\{u\leq t\}}}{1-H(u|x)},\, t\in[\tau_0,\tau_1]\right\}\\
&=:&\mathcal{F}_n+\mathcal{G}_{n,1}+\mathcal{G}_{n,2}.
\end{eqnarray*}
According to Lemma 2.6.18 {\it(i)}, {\it(vi)} and {\it(viii)} in \cite{Vaart1996}, we obtain that $\mathcal{G}_{n,1}$ and $\mathcal{G}_{n,2}$ are \textit{VC} with the envelope function $E'_n(u,v,v',w)=\sqrt{b^p}K_b(x-w)(1-H(\tau_1|x))^{-1}$. Finally, since the covering number for $\mathcal{F}_n$ is already provided in \cite{Escobar2019} and that $\widehat{\mathcal{F}}_n$ is included in the class of functions $\mathcal{F}_n+\mathcal{G}_{n,1}+\mathcal{G}_{n,2}$ with envelope function $E_n+2E'_n$, we have from the Lemma 16 in \cite{Nolan1987}, for any $t>0$ 
\begin{eqnarray*}
\sup_\mathcal{Q} N(\widehat{\mathcal{F}}_n,L_2(Q),t\Vert E_n+2E'_n\Vert_{Q,2})&\leq&\sup_\mathcal{Q} N(\mathcal{F}_n+\mathcal{G}_{n,1}+\mathcal{G}_{n,2},L_2(Q),t\Vert E_n+2E'_n\Vert_{Q,2})\\
&\leq& L\left(\dfrac{1}{t}\right)^V
\end{eqnarray*}
for some $L$ and $V$. Thus, (\ref{ST2}) is established since for any sequence $\delta_n\searrow 0$ and $n$ large enough
\begin{eqnarray*}
J(\delta_n,\widehat{\mathcal{F}}_n,L_2)
&\leq&\int_0^{\delta_n}\sqrt{\log(2^VL)-V\log(t)}dt=o(1).
\end{eqnarray*}
and the weak convergence for our process is established. In order to derive the covariance structure of the limiting process, it is sufficient to see that in ($\ref{proof_g}$), $\widehat{g}$ might be written as
\begin{eqnarray*}
\widehat{g}(t,T,\delta,P|x)=(1-F(t|x))\left\{\dfrac{P\ind_{\{T\leq t\}}}{1-H(T|x)}-\int_{-\infty}^{T\nonumber\wedge t}\dfrac{dH^u(y|x)}{(1-H(y|x))^2}\right\}.
\end{eqnarray*}
with the following equalities 
\begin{eqnarray*}
\mathbb{E}\left[\left.\dfrac{P\ind_{\{T\leq t\}}}{1-H(T|x)}\int_{-\infty}^{T\wedge s}\dfrac{dH^u(y|x)}{(1-H(y|x))^2}\right| X=x\right]&=&\mathbb{E}\left[\left.\dfrac{\ind_{\{\delta=1, T\leq t\}}}{1-H(T|x)}\int_{-\infty}^{T\wedge s}\dfrac{dH^u(y|x)}{(1-H(y|x))^2}\right| X=x\right]\\
&=&\int_{-\infty}^t\dfrac{1}{1-H(z|x)}\int_{-\infty}^{z\wedge s}\dfrac{dH^u(y|x)}{(1-H(y|x))^2}dH^u(z|x),
\end{eqnarray*}
\begin{eqnarray*}
\hspace{-3.8cm}\mathbb{E}\left[\left.\dfrac{P^2\ind_{\{T\leq t,T\leq s\}}}{(1-H(T|x))^2}\right| X=x\right]=\int_{-\infty}^{t\wedge s}\dfrac{\mathbb{E}[P^2|T=y,X=x]}{(1-H(y|x))^2} dH(y|x)\\
\end{eqnarray*}
and
\begin{eqnarray*}
&&\hspace{-.9cm}\mathbb{E}\left[\left.\int_{-\infty}^{T\wedge t}\dfrac{dH^u(y|x)}{(1-H(y|x))^2}\int_{-\infty}^{T\wedge s}\dfrac{dH^u(y|x)}{(1-H(y|x))^2}\right| X=x\right]\\
&=&\int_{-\infty}^{+\infty}\int_{-\infty}^{z\wedge t}\dfrac{dH^u(y|x)}{(1-H(y|x))^2}\int_{-\infty}^{z\wedge s}\dfrac{dH^u(y|x)}{(1-H(y|x))^2}dH(z|x)\\
&=&\int_{-\infty}^t\dfrac{1}{1-H(z|x)}\int_{-\infty}^{z\wedge s}\dfrac{dH^u(y|x)}{(1-H(y|x))^2}dH^u(z|x)+\int_{-\infty}^s\dfrac{1}{1-H(z|x)}\int_{-\infty}^{z\wedge t}\dfrac{dH^u(y|x)}{(1-H(y|x))^2}dH^u(z|x).\\
\end{eqnarray*}
We establish the continuity of the process thanks to a sufficient condition due to \cite{Fernique1964}. Indeed, let $(s,t)\in\mathbb{R}^2$ and denote 
\begin{eqnarray*}
\widebar{F}(\cdot|x)=1-F(\cdot|x)\quad\text{and}\quad m(y|x)=\dfrac{\mathbb{E}[P^2|T=y,X=x]}{(1-H(y|x))^2}.
\end{eqnarray*}
Then,
\begin{eqnarray*}
&&\mathbb{E}[(Z(s|x)-Z(t|x))^2]\dfrac{f(x)}{\Vert K\Vert ^2_2}\\
&&=\widebar{F}(s|x)^2\int_{-\infty}^{s}m(y|x)dH(y|x) -2\widebar{F}(s|x)\widebar{F}(t|x)\int_{-\infty}^{s\wedge t}m(y|x)dH(y|x)+\widebar{F}(t|x)^2\int_{-\infty}^{t}m(y|x)dH(y|x)\\
&&=\widebar{F}(s|x)\left[\widebar{F}(s|x)\int_{s\wedge t}^{s}m(y|x)dH(y|x)+(F(t|x)-F(s|x))\int_{-\infty}^{s\wedge t}m(y|x)dH(y|x)\right]\\
&& \quad +\widebar{F}(t|x)\left[\widebar{F}(t|x)\int_{s\wedge t}^{t}m(y|x)dH(y|x)+(F(s|x)-F(t|x))\int_{-\infty}^{s\wedge t}m(y|x)dH(y|x)\right]\\
&&=\widebar{F}(s\vee t|x)^2\int_{s\wedge t}^{s\vee t}m(y|x)dH(y|x)+\int_{-\infty}^{s\wedge t}m(y|x)dH(y|x)(F(s|x)-F(t|x))^2\\
&&\leq \dfrac{c|s-t|^{\eta'}}{(1-H(\tau_1|x))^2}.
\end{eqnarray*}
This yields that $\sqrt{\mathbb{E}[(Z(s|x)-Z(t|x))^2]}\leq\xi(t-s)$, with 
$$ \xi(t-s)=\sqrt{\dfrac{c}{f(x)}}\dfrac{\Vert K\Vert_2|s-t|^{\eta'/2}}{1-H(\tau_1|x)} $$ 
where $\xi$ is monotone and
\begin{eqnarray*}
\int_0^1\dfrac{\xi(u)}{u|\log(u)|^{1/2}}du<+\infty.\\ 
\end{eqnarray*}
Finally, the last assertion is a direct consequence from the Jensen's inequality.\hfill$\Box$\\

\bigskip
\noindent
\textbf{Proof of the Proposition \ref{prop:plugin}}.\\ 
According to the proof of Lemma \ref{lambda}, we also have that 
\begin{eqnarray*}
\sup_{t\in[\tau_0,\tau_1]}\sqrt{nh^p}\left|1-\widecheck{F}_n(t|x)-\exp(-\widecheck{\Lambda}_n(t|x))\right|=o_\mathbb{P}(1)
\end{eqnarray*}  
where $\widecheck{\Lambda}$ defines the cumulative hazard function given by
\begin{eqnarray*}
\widecheck{\Lambda}_n(t|x):=\int_{-\infty}^t\dfrac{d\widecheck{H}_n^u(s|x)}{1-H_n(s^-|x)}\quad\text{with}\quad\widecheck{H}_n^u(t|x):=\sum_{i=1}^nW_b(x-X_i)\ind_{\{T_i\leq t\}}p_n(T_i,X_i).
\end{eqnarray*}
Using the integration by parts formula for Stieltjes integrals, it follows that
\begin{eqnarray*}
\widecheck{\Lambda}_n(t|x)&=&\widehat{\Lambda}_n(t|x)+\int_{-\infty}^t\dfrac{d(\widecheck{H}_n^u-\widehat{H}_n^u)(s|x)}{1-H_n(s^-|x)}\\
&=&\widehat{\Lambda}_n(t|x)+\dfrac{\widecheck{H}_n^u(t|x)-\widehat{H}_n^u(t|x)}{1-H_n(t|x)}-\int_{-\infty}^t\widecheck{H}_n^u(s|x)-\widehat{H}_n^u(s|x)d\left(\dfrac{1}{1-H_n}\right)(s|x)\\
&=&\widehat{\Lambda}_n(t|x)+\dfrac{\widecheck{H}_n^u(t|x)-\widehat{H}_n^u(t|x)}{1-H_n(t|x)}-\int_{-\infty}^t\dfrac{\widecheck{H}_n^u(s|x)-\widehat{H}_n^u(s|x)}{(1-H_n(s|x))^2}dH_n(s|x).
\end{eqnarray*}
Since $H_n$ is a non-decreasing function, this yields 
\begin{eqnarray*}
\sup_{t\in[\tau_0,\tau_1]}\left|\widecheck{\Lambda}_n(t|x)-\widehat{\Lambda}_n(t|x)\right|=O_\mathbb{P}\left(\sup_{t\in[\tau_0,\tau_1]}\left|\widecheck{H}_n^u(t|x)-\widehat{H}_n^u(t|x)\right|\right)
\end{eqnarray*}
where
\begin{eqnarray*}
\sup_{t\in[\tau_0,\tau_1]}\left|\widecheck{H}_n^u(t|x)-\widehat{H}_n^u(t|x)\right|\leqslant\sum_{i=1}^nW_b(x-X_i)\left|p_n(T_i,X_i)-p(T_i,X_i)\right|.
\end{eqnarray*}
By Markov's and Hölder's inequalities, this immediately gives $\forall\varepsilon>0$
\begin{eqnarray*}
&&\hspace{-2cm}\mathbb{P}\left((nb^p)^{-1/2}\sup_{t\in[\tau_0,\tau_1]}\left|\widecheck{H}_n^u(t|x)-\widehat{H}_n^u(t|x)\right|>\varepsilon\right)\\
&&\leqslant\varepsilon^{-1}(nb^p)^{-1/2}\mathbb{E}\left[W_h(x-X)\left|p_n(T,X)-p(T,X)\right|\right]\\
&&\leqslant\varepsilon^{-1}(nb^p)^{-1/2}\mathbb{E}\left[\left|p_n(T,X)-p(T,X)\right|^q\right]^{1/q}\left(\Vert K\Vert_\infty^{(q-1)/q}+o_\mathbb{P}(1)\right)
\end{eqnarray*}
and the result follows. \hfill$\Box$\\

\bigskip
\noindent
\textbf{Proof of the Theorem \ref{theorem:vc}}.\\ 
The results are mainly consequences from Corollary 3.2.3, Theorem 3.2.5 and various straightforward properties for \textit{VC} classes from the section 2.6 in \cite{Vaart1996}. The same tools from the proof of Proposition 5 are also used, aside that $Q_\bold{x}$ and  $Q_\bold{y}$ respectively define the law of the vectors $\bold{X}=(T,\xi,X)$ and $\bold{Y}=(T,\xi,\delta,X)$. All along, we consider the metric spaces $(\mathcal{L},d,Q_\bold{x})$ and $(\mathcal{H},d,Q_\bold{y})$ where $d$ denotes the embedded $L_2$ distance.\\ 
Without lost of generality, we can work with uniform margins $D_i$ by replacing the class of functions $\mathcal{H}$ and $\mathcal{L}$ with $C_1(\mathcal{H})$ and $D_2(\mathcal{L})$ respectively. Since the $D_i$'s are monotone functions, the latter class of functions are still $\textit{VC}$ with functions uniformly bounded away from zero. Mostly, one can retrieve the initial estimators $\widehat{h}_n$ and $\widehat{\ell}_n$ by use of the inverse transformations $D^{-1}_i$.\\

\noindent
\textbf{Step 1} :   We first prove the $L_2$ consistency for $\widehat{\ell}_n$. Let us define the class of functions
\begin{eqnarray*}
\mathcal{M}_\mathcal{L}=\left\{m_\ell(t,b,x)= b\log[\ell(t,x)]+(1-b)\log[1-\ell(t,x)],\, \ell\in\mathcal{L}\right\}.
\end{eqnarray*}
Because  the function $\log$ is monotone, the \textit{VC} property for the class of functions $\mathcal{L}$ is also true for $\mathcal{M}_\mathcal{L}$. According to Theorem 2.8.1 in \cite{Vaart1996}, it follows that $\mathcal{M}_\mathcal{L}$ is actually Glivenko-Cantelli when only considering the measure $Q$, meaning that
\begin{eqnarray*}
\label{Gliv_l}
\mathbb{E}\left[\sup_{\ell\in\mathcal{L}}\left\vert \mathbb{M}^{(2)}_n(\ell)-Q_\bold{x}m_\ell\right\vert\right]\longrightarrow 0,\quad\text{as }n\to+\infty.
\end{eqnarray*}
The strict concavity of the logarithmic function implies that $Q_\bold{x}m_{\ell_0}\geqslant Q_\bold{x}m_\ell$ for any $\ell\in\mathcal{L}$, whom by use of the Taylor's expansion, gives
\begin{eqnarray*}
Q_\bold{x}m_\ell-Q_\bold{x}m_{\ell_0}&\leq&-\dfrac{Q_\bold{x}(\ell- \ell_0)^2}{2(r_2\wedge(1-r_2))^2}=-\dfrac{d(\ell, \ell_0)^2}{2(r_2\wedge(1-r_2))^2}.
\end{eqnarray*}
This allows us to apply Corollary 3.2.3 and proves that $\forall\varepsilon>0$,  $\mathbb{P}(d(\widehat{\ell}_n,\ell_0)>\varepsilon)\to 0$. We furthermore show that the convergence in probability can be strengthened, in the sense that
\begin{eqnarray}
\label{proof:rn2}
\mathbb{P}(q_{n,2}d(\widehat{\ell}_n,\ell_0)>\varepsilon)\to 0,\quad \forall\varepsilon>0
\end{eqnarray}
where $q_{n,2}^2\psi_n^{(2)}(1/q_{n,2})\leq\sqrt{n}$ for $n$ large enough. Indeed, let us consider the class of functions 
\begin{eqnarray*}
\mathcal{M}_{\mathcal{L},\varepsilon}=\left\{m_\ell-m_{\ell_0},\,\ell\in\mathcal{L}\text{ and } d(\ell,\ell_0)\leqslant\varepsilon\right\},\quad\varepsilon\geqslant0.
\end{eqnarray*}
Clearly, $\mathcal{M}_{\mathcal{L},\varepsilon}$ is \textit{VC} independently from $\varepsilon\geqslant0$ since $\mathcal{M}_\mathcal{L}$ is too. According to Theorem 2.1 in \cite{Gine2002}, for $\sigma^2\geq\sup_{m\in\mathcal{M}_{\mathcal{L},\varepsilon}}\text{Var}(m)$, $U_2\geq \sup_{m\in\mathcal{M}_{\mathcal{L},\varepsilon}}\Vert m \Vert_\infty$ and $0<\sigma_2\leq U_2$, there exists some universal constants $A,B$ such that   
\begin{eqnarray*}
&&\mathbb{E}\left[\displaystyle\sup_{d(\ell,\ell_0)\leqslant\varepsilon}\left\vert \mathbb{M}^{(2)}_n(\ell)-Q_\bold{x}m_\ell-(\mathbb{M}^{(2)}_n(\ell_0)-Q_\bold{x}m_{\ell_0})\right\vert\right]\\
&=&\mathbb{E}\left[\displaystyle\sup_{d(\ell,\ell_0)\leqslant\varepsilon}\left\vert \dfrac{1}{n}\displaystyle\sum_{i=1}^n(m_\ell-m_{\ell_0})(\bold{X}_i)-Q_\bold{x}(m_\ell-m_{\ell_0})\right\vert\right]
\leq B\left[ \frac{U_2\nu}{n}\log\left(\dfrac{U_2A}{\sigma_2}\right)+\sqrt{\frac{\nu\sigma_2^2}{n}\log\left(\dfrac{U_2A}{\sigma_2}\right) }\,\right]
\end{eqnarray*}
where $\nu$ is the \textit{VC} index associated to $\mathcal{M}_{\mathcal{L}}$. Here, we can choose
\begin{eqnarray*}
U_2\geq\log\left(\dfrac{R_2}{r_2}\right)\vee\log\left(\dfrac{1-r_2}{1-R_2}\right)\quad\text{and}\quad\sigma_2^2=\varepsilon^2\left(\dfrac{R_2}{r_2^2}\vee\dfrac{1-r_2}{(1-R_2)^2}\right).
\end{eqnarray*}
This particularly shows that for $\varepsilon$ small enough, $\sigma\leqslant U$ and
\begin{eqnarray}
\label{proof:m2}
\mathbb{E}\left[\displaystyle\sup_{d(\ell,\ell_0)\leqslant\varepsilon}\left\vert \mathbb{M}^{(2)}_n(\ell)-Q_\bold{x}m_\ell-(\mathbb{M}^{(2)}_n(\ell_0)-Q_\bold{x}m_{\ell_0})\right\vert\right]\leq \widetilde{R}\dfrac{\psi^{(2)}_n(\varepsilon)}{\sqrt{n}}
\end{eqnarray}
where $\widetilde{R}$ is some universal constant independent from $n$ and $\varepsilon$. The result for the first step hence directly follows from Theorem 3.2.5 in \cite{Vaart1996}.\\ 

\bigskip
\noindent
\textbf{Step 1.2} : The forthcoming proofs mostly rely on the asymptotic behavior of $\widehat{\ell}_n$, so that one expect replacing the estimator with $\ell_0$ with almost no lost of efficiency. However, the dependence between $\widehat{\ell}_n$ and the initial sample might contradict the equality between $\mathbb{E}[d(\widehat{\ell}_n,\ell_0)]$ and $\mathbb{E}[(\widehat{\ell}_n-\ell_0)^2(T_1,X_1)]$, meaning that (\ref{proof:rn2}) is not a guarantee for the small discrepancy between $\widehat{\ell}_n$ and $\ell_0$ when applied to $\mathbb{M}^{(1)}_n$. We will thus show that we can conserve the same speed of convergence than that of (\ref{proof:rn2}) and prove the first assertion in (\ref{proof:speed}). Let us define the classes of functions
\begin{eqnarray*}
&&\dot{\mathcal{M}}_\mathcal{L}=\left\{\dot{m}_\ell(t,b,x)= \ell(t,x)\left[\dfrac{b}{\ell_0(t,x)}-\dfrac{1-b}{1-\ell_0(t,x)}\right],\,\ell\in\mathcal{L}\right\}\\
\text{and}&&\\
&&\dot{\mathcal{M}}_{\mathcal{L},\varepsilon}=\left\{\dot{m}_\ell-\dot{m}_{\ell_0},\,\ell\in\mathcal{L}\text{ and } d(\ell,\ell_0)\leqslant\varepsilon\right\},\quad\varepsilon\geqslant0.
\end{eqnarray*}
Likewise $\mathcal{M}_\mathcal{L}$ and $\mathcal{M}_{\mathcal{L},\varepsilon}$, the classes $\dot{\mathcal{M}}_\mathcal{L}$ and $\dot{\mathcal{M}}_{\mathcal{L},\varepsilon}$ are \textit{VC} independently from $\varepsilon\geq 0$. Applying once again the Theorem 2.1 in \cite{Gine2002} with this time 
\begin{eqnarray*}
\dot{U}_2\geq\dfrac{2R_2(1-R_2)}{r_2(1-r_2)}\quad\text{and}\quad\dot{\sigma}_2=\dfrac{\varepsilon^2}{r_2(1-R_2)}.
\end{eqnarray*}
we obtain
\begin{eqnarray}
\label{proof:mdot2}
\mathbb{E}\left[\displaystyle\sup_{d(\ell,\ell_0)\leqslant\varepsilon}\left\vert \dot{\mathbb{M}}^{(2)}_n(\ell)-Q_\bold{x}\dot{m}_\ell-(\dot{\mathbb{M}}^{(2)}_n(\ell_0)-Q_\bold{x}\dot{m}_{\ell_0})\right\vert\right]\leq \widebar{R}\dfrac{\psi^{(2)}_n(\varepsilon)}{\sqrt{n}}
\end{eqnarray}
where 
\begin{eqnarray*}
\dot{\mathbb{M}}^{(2)}_n(\ell)=\dfrac{1}{n}\sum_{i=1}^n\dot{m}_\ell(\bold{X}_i)=\dfrac{1}{n}\sum_{i=1}^n\ell(T_i,X_i)\left[\dfrac{\xi_i}{\ell_0(T_i,X_i)}-\dfrac{1-\xi_i}{\ell_0(T_i,X_i)}\right]
\end{eqnarray*}
 and $\widebar{R}$ is some another universal constant independent from $n$ and $\varepsilon$. By Taylor's expansion up to the second order for the logarithmic function, we have 
\begin{eqnarray*}
&&\mathbb{M}^{(2)}_n(\widehat{\ell}_n)-\mathbb{M}^{(2)}_n(\ell_0)\\
&=&\dot{\mathbb{M}}^{(2)}_n(\widehat{\ell}_n)-\dot{\mathbb{M}}^{(2)}_n(\ell_0)-\dfrac{1}{n}\sum_{i=1}^n\dfrac{1}{2}(\widehat{\ell}_n-\ell_0)^2(T_i,X_i)\left(\dfrac{\xi_i}{\widetilde{\ell}_{1,n}(T_i,X_i)^2}+\dfrac{1-\xi_i}{(1-\widetilde{\ell}_{2,n}(T_i,X_i)^2)}\right)
\end{eqnarray*}
where $\widetilde{\ell}_{j,n}(T_i,X_i)$, $j=1,2$ are random values ranging between $\widehat{\ell}_n(T_i,X_i)$ and $\ell_0(T_i,X_i)$ almost surely. A straightforward reformulation shows that almost surely
\begin{eqnarray*}
&&\frac{1}{2(m_2\wedge(1-M_2))^2}\dfrac{1}{n}\sum_{i=1}^n(\widehat{\ell}_n-\ell_0)^2(T_i,X_i)\\
&\leq&\left|\mathbb{M}^{(2)}_n(\widehat{\ell}_n)-Q_\bold{x}m_{\widehat{\ell}_n}-(\mathbb{M}^{(2)}_n(\ell_0)-Q_\bold{x}m_{\ell_0})\right|+\left|\dot{\mathbb{M}}^{(2)}_n(\widehat{\ell}_n)-Q_\bold{x}\dot{m}_{\widehat{\ell}_n}-(\dot{\mathbb{M}}^{(2)}_n(\ell_0)-Q_\bold{x}\dot{m}_{\ell_0})\right|\\
&+&|Q_\bold{x}m_{\widehat{\ell}_n}-Q_\bold{x}m_{\ell_0}-Q_\bold{x}\dot{m}_{\widehat{\ell}_n}+Q_\bold{x}\dot{m}_{\ell_0}|.
\end{eqnarray*}
Note that the previous choices of $U_2$ and $\dot{U}_2$ can be arbitrary large such that $\sigma_2\leq U_2$ and $\dot{\sigma}_2\leq\dot{U}_2$, but $\varepsilon$ is large enough to have $d(\ell,\ell_0)\leq\varepsilon$ for any $\ell\in\mathcal{L}$ and the properties (\ref{proof:m2}) and (\ref{proof:mdot2}) are still valid (for possibly larger constants $\widetilde{R}$ and $\widebar{R}$). Hence, this implies that
\begin{eqnarray}
\label{proof:L2distance}
\nonumber\mathbb{E}\left[\dfrac{1}{n}\sum_{i=1}^n(\widehat{\ell}_n-\ell_0)^2(T_i,X_i)\right]\lesssim\mathbb{E}\left[|Q_\bold{x}(m_{\widehat{\ell}_n}-m_{\ell_0}-\dot{m}_{\widehat{\ell}_n}+\dot{m}_{\ell_0})|\right] + O(n^{-1/2}).
\end{eqnarray}
Similarly, one can extend the function inside $Q_\bold{x}$, which almost surely gives 
\begin{eqnarray*}
|Q_\bold{x}(m_{\widehat{\ell}_n}-m_{\ell_0}-\dot{m}_{\widehat{\ell}_n}+\dot{m}_{\ell_0})|\leq\dfrac{1}{2(r_2\wedge(1-R_2))^2}d(\widehat{\ell}_n,\ell_0)^2
\end{eqnarray*}
and proves the first part of (\ref{proof:speed}).\\ 

\bigskip
\noindent
\textbf{Step 2.1} : We next show the $L_2$ consistency for $\widehat{h}_n$ by use of the same approach than that of the step 1. Let us now define the class of functions $\mathcal{M}_\mathcal{H}$ given by
\begin{eqnarray*}
\mathcal{M}_\mathcal{H}=\left\{m_{h,\ell_0}(t,a,b,x)= ab\log[C(h,\ell_0)](x,t)+a(1-b)\log[\ell_0-C(h,\ell_0)](x,t),\, h\in\mathcal{H}\right\}.
\end{eqnarray*}
With $D$ monotone in both arguments, one can show that $\mathcal{M}_\mathcal{H}$ is also \textit{VC} with a small adaption of the proof of Lemma 2.6.18 (viii) (it is sufficient to replace $\phi$ by $D(.\,,\ell_0)$ and to use the monotony behavior of $D$). Likewise (\ref{Gliv_l}), $\mathcal{M}_\mathcal{H}$ is Glivenko-Cantelli and by extension we obtain that
\begin{eqnarray*}
&&\mathbb{E}\left[\sup_{h\in\mathcal{H}}\left\vert \mathbb{M}^{(1)}_n(h,\widehat{\ell}_n)-Q_\bold{y}m_{h,\ell_0}\right\vert\right]\\
&\leq&\mathbb{E}\left[\sup_{h\in\mathcal{H}}\left\vert\mathbb{M}^{(1)}_n(h,\widehat{\ell}_n)-\mathbb{M}^{(1)}_n(h,\ell_0)\right\vert\right]+\mathbb{E}\left[\sup_{h\in\mathcal{H}}\left\vert \mathbb{M}^{(1)}_n(h,\ell_0)-Q_\bold{y}m_{h,\ell_0}\right\vert\right]\\
&\lesssim&\mathbb{E}\left[\dfrac{1}{n}\sum_{i=1}^n(\widehat{\ell}_n-\ell_0)^2(T_i,X_i)\right]+o(1)
\end{eqnarray*}
where the left term in the last inequality results once again from the the Cauchy-Schwarz inequality and the differentiability for both the $\log$ function and $D$. Likewise $\ell_0$, we have to show that $h_0$ actually defines a strict minimum for $Q_\bold{y}m_{h,\ell_0}$. This time, we obtain from the Taylor's expansion that
\begin{eqnarray*}
Q_\bold{y}m_{h,\ell_0}-Q_\bold{y}m_{h_0,\ell_0}&\leq&-\dfrac{Q_\bold{y}\left[(D(h,\ell_0)-D(h_0,\ell_0))^2\right]}{D(r_1,r_2)^2}\leqslant-\dfrac{k^2}{D(r_1,r_2)^2}d(h,h_0)^2
\end{eqnarray*}
and the same conclusion is drawn from Corollary 3.2.3 with $\forall\varepsilon>0$, $\mathbb{P}(d(\widehat{h}_n,h_0)>\varepsilon)\to 0$ as $n\to+\infty$. The speed of convergence is also obtained with the same previous approach. Define the \textit{VC} class of functions
\begin{eqnarray*}
\mathcal{M}_{\mathcal{H},\varepsilon}=\left\{m_{h,\ell_0}-m_{h_0,\ell_0},\,h\in\mathcal{H}\text{ and } d(h,h_0)\leqslant\varepsilon\right\},\quad\varepsilon\geqslant0.
\end{eqnarray*}
Applying once again Proposition 2.1 in \cite{Gine2002} with 
\begin{eqnarray*}
U_1=\sup_{h\in\mathcal{H}}\max\left\{\log\left(\dfrac{D(h,\ell_0)}{D(h_0,\ell_0)}\right),\log\left(\dfrac{1-D(h,\ell_0)}{1-D(h_0,\ell_0)}\right)\right\}\quad\text{and}\quad\sigma_1=R^2\dfrac{\varepsilon^2}{D(r_1,r_2)\wedge k(1-R_2)}.
\end{eqnarray*}
we have for $\varepsilon$ small enough, $\sigma\leqslant U$ and
\begin{eqnarray}
\label{proof:m1}
\mathbb{E}\left[\displaystyle\sup_{d(h,h_0)\leqslant\varepsilon}\left\vert \mathbb{M}^{(1)}_n(h,\ell_0)-Q_\bold{y}m_{h,\ell_0}-\mathbb{M}^{(1)}_n(h_0,\ell_0)+Q_\bold{y}m_{h_0,\ell_0}\right\vert\right]\leq \widebar{R}\dfrac{\psi^{(2)}_n(\varepsilon)}{\sqrt{n}}
\end{eqnarray}
where $\widebar{R}$ is another universal constant independent from $n$ and $\varepsilon$.\\ 

\bigskip
\noindent
\textbf{Step 2.2} : To complete the proof, we have to show that $\ell_0$ can be replaced by $\widehat{\ell}_n$ in (\ref{proof:m1}), i.e. we want
\begin{eqnarray}
\label{proof:m22}
\mathbb{E}\left[\displaystyle\sup_{d(h,h_0)\leqslant\varepsilon}\left\vert \dfrac{1}{n}\sum_{i=1}^n[m_{h,\ell_0}-m_{h,\widehat{\ell}_n}-m_{h_0,\ell_0}+m_{h_0,\widehat{\ell}_n}](\bold{Y}_i)\right\vert\right]\lesssim\dfrac{\psi^{(1)}_n(\varepsilon)}{\sqrt{n}}
\end{eqnarray}
since $\frac{\psi^{(2)}_n(\varepsilon)}{\psi^{(1)}_n(\varepsilon)}\leq 1$ when $n$ is large enough and $\varepsilon$ is small enough. By assumptions, the function $(u_1,u_2)\to\log(D(u_1,u_2))$ is well defined and differentiable such that for any $(t,x)\in\mathbb{R}_+\times\mathbb{R}^p$ and $h\in\mathcal{H}$
\begin{eqnarray*}
\left[\log(\ell_0-D(h,\ell_0))-\log(\widehat{\ell}_n-D(h,\widehat{\ell}_n))\right](t,x)&=&\left[\int_{\widehat{\ell}_n}
^{\ell_0}\left(1-\dfrac{\partial D}{\partial u_2}(h,s)\right)\left(s-D(h,s)\right)^{-1}ds\right](t,x)\\
\text{and}\quad\left[\log(D(h,\ell_0))-\log(D(h,\widehat{\ell}_n))\right](t,x)&=&\left[\int_{\widehat{\ell}_n}
^{\ell_0}\dfrac{\partial D}{\partial u_2}(h,s)D(h,s)^{-1}ds\right](t,x).
\end{eqnarray*}
According to the uniform Lipschitz property for the derivative function, we have that both the equations
\begin{eqnarray*}
&&\left[\int_{\ell_0\wedge\widehat{\ell}_n}
^{\ell_0\vee\widehat{\ell}_n}\left|\left(1-\dfrac{\partial D}{\partial u_2}(h,s)\right)(s-D(h,s))^{-1}-\left(1-\dfrac{\partial D}{\partial u_2}(h_0,s)\right)(s-D(h_0,s))^{-1}\right|ds\right](t,x)\\
&\text{and}&\quad\left[\int_{\ell_0\wedge\widehat{\ell}_n}
^{\ell_0\vee\widehat{\ell}_n}\left|\dfrac{\partial D}{\partial u_2}(h,s)D(h,s)^{-1}-\dfrac{\partial D}{\partial u_2}(h_0,s)D(h_0,s)^{-1}\right|ds\right](t,x)
\end{eqnarray*}
are bounded up to a uniform constant by $|h-h_0||\widehat{\ell}_n-\ell_0|(t,x)$. Since we have for any $a,b\in\{0,1\}$
\begin{eqnarray*}
&&[m_{h,\ell_0}-m_{h,\widehat{\ell}_n}-m_{h_0,\ell_0}+m_{h_0,\widehat{\ell}_n}](t,a,b,x)\\
&\leq&\left[\left|\log(D(h,\ell_0))-\log( D(h,\widehat{\ell}_n))+\log(D(h_0,\widehat{\ell}_n))-\log(D(h_0,\ell_0))\right|\right](t,x)\\
&+&\left[\left|\log(\ell_0-D(h,\ell_0))-\log(\widehat{\ell}_n- D(h,\widehat{\ell}_n))+\log(\widehat{\ell}_n-D(h_0,\widehat{\ell}_n))-\log(\ell_0-D(h_0,\ell_0))\right|\right](t,x)
\end{eqnarray*}
these imply together with the aforementioned property and the Cauchy-Schwarz inequality that
\begin{eqnarray*}
&&\mathbb{E}\left[\displaystyle\sup_{d(h,h_0)\leqslant\varepsilon}\left\vert \dfrac{1}{n}\sum_{i=1}^n[m_{h,\ell_0}-m_{h,\widehat{\ell}_n}-m_{h_0,\ell_0}+m_{h_0,\widehat{\ell}_n}](\bold{Y}_i)\right\vert\right]\\
&\lesssim& \mathbb{E}\left[\sup_{d(h,h_0)\leqslant\varepsilon} \dfrac{1}{n}\sum_{i=1}^n|h-h_0|\times|\widehat{\ell}_n-\ell_0|(T_i,X_i)\right]\\
&\leq&\mathbb{E}\left[\sup_{d(h,h_0)\leqslant\varepsilon} \dfrac{1}{n}\sum_{i=1}^n(h-h_0)^2(T_i,X_i)\right]^{1/2}\mathbb{E}\left[\dfrac{1}{n}\sum_{i=1}^n(\widehat{\ell}_n-\ell_0)^2(T_i,X_i)\right]^{1/2}.
\end{eqnarray*}
Lastly, we have that the class of functions $(h-h_0)^2$ where $h$ ranges in $\mathcal{H}$ is \textit{VC}. The same arguments used all along this proof allows us to show that
\begin{eqnarray*}
\mathbb{E}\left[\sup_{d(h,h_0)\leqslant\varepsilon} \dfrac{1}{n}\left|\sum_{i=1}^n(h-h_0)^2(T_i,X_i)-d(h,h_0)^2\right|\right]\lesssim\dfrac{\psi^{(2)}_n(\varepsilon)}{\sqrt{n}}
\end{eqnarray*}
and thus
\begin{eqnarray*}
\mathbb{E}\left[\sup_{d(h,h_0)\leqslant\varepsilon} \dfrac{1}{n}\sum_{i=1}^n(h-h_0)^2(T_i,X_i)\right]\lesssim\dfrac{\psi^{(2)}_n(\varepsilon)}{\sqrt{n}}+\varepsilon^2.
\end{eqnarray*}
By use of the first part of (\ref{proof:speed}) together with the latter inequalities, these conclude the proof of (\ref{proof:m22}) since $q_{n,1}=O(\sqrt{n})$. Combining (\ref{proof:m1}) and (\ref{proof:m22}), we have 
\begin{eqnarray*}
\label{proof:m3}
\mathbb{E}\left[\displaystyle\sup_{d(h,h_0)\leqslant\varepsilon}\left\vert \mathbb{M}^{(1)}_n(h,\widehat{\ell}_n)-Q_\bold{y}m_{h,\ell_0}-(\mathbb{M}^{(1)}_n(h_0,\widehat{\ell}_n)-Q_\bold{y}m_{h_0,\ell_0})\right\vert\right]\leq \widebar{K}\dfrac{\psi^{(1)}_n(\varepsilon)}{\sqrt{n}}
\end{eqnarray*}
which again implies from Theorem 3.2.5 in \cite{Vaart1996} that  
\begin{eqnarray*}
\label{proof:rn1}
\mathbb{P}(r_{n,1}d(\widehat{h}_n,h_0)>\varepsilon)\to 0,\quad \forall\varepsilon>0.
\end{eqnarray*}

\bigskip
\noindent
\textbf{Step 2.3} : It finally remains to prove the second part in (\ref{proof:speed}). Likewise $\widehat{\ell}_n$ in step 1.2, we will use the Taylor's expansion up to the second order for the function $m_{h,\ell_0}$ in $h$. Let us define the classes of functions
\begin{eqnarray*}
&&\dot{\mathcal{M}}_\mathcal{H}=\left\{\dot{m}_{h,\ell_0}(t,a,b,x)= \left[D(h,\ell_0)\times\left(\dfrac{ab}{D(h_0,\ell_0)}-\dfrac{a(1-b)}{\ell_0-D(h_0,\ell_0)}\right)\right](t,x),\,h\in\mathcal{H}\right\}\\
\text{and}&&\\
&&\dot{\mathcal{M}}_{\mathcal{H},\varepsilon}=\left\{\dot{m}_{h,\ell_0}-\dot{m}_{h_0,\ell_0},\,\ell\in\mathcal{H}\text{ and } d(h,h_0)\leqslant\varepsilon\right\},\quad\varepsilon\geqslant0.
\end{eqnarray*}
Likewise $\mathcal{M}_\mathcal{H}$ and $\mathcal{M}_{\mathcal{H},\varepsilon}$, the classes $\dot{\mathcal{M}}_\mathcal{H}$ and $\dot{\mathcal{M}}_{\mathcal{H},\varepsilon}$ are \textit{VC} independently from $\varepsilon\geq 0$. By use of the same arguments than that of the step 1.2, we obtain
\begin{eqnarray}
\label{proof:mdot1}
\mathbb{E}\left[\displaystyle\sup_{d(h,h_0)\leqslant\varepsilon}\left\vert \dot{\mathbb{M}}^{(1)}_n(h,\ell_0)-Q_\bold{y}\dot{m}_{h,\ell_0}-\left(\dot{\mathbb{M}}^{(1)}_n(h_0
,\ell_0)-Q_\bold{y}\dot{m}_{h_0,\ell_0}\right)\right\vert\right]\leq \widecheck{R}\dfrac{\psi^{(2)}_n(\varepsilon)}{\sqrt{n}}
\end{eqnarray}
where 
\begin{eqnarray*}
\dot{\mathbb{M}}^{(1)}_n(h,\ell_0)=\dfrac{1}{n}\sum_{i=1}^n\dot{m}_{h,\ell_0}(\bold{Y}_i)=\dfrac{1}{n}\sum_{i=1}^n\left[D(h,\ell_0)\times\left(\dfrac{\xi_i\delta_i}{D(h_0,\ell_0)}-\dfrac{\xi_i(1-\delta_i)}{\ell_0-D(h_0,\ell_0)}\right)\right](T_i,X_i)
\end{eqnarray*}
 and $\widecheck{R}$ is some another universal constant independent from $n$ and $\varepsilon$. These allows us to write
 \begin{eqnarray*}
 &&\mathbb{M}^{(1)}_n(\widehat{h}_n,\ell_0)-\mathbb{M}^{(1)}_n(h_0,\ell_0)-\dot{\mathbb{M}}^{(1)}_n(\widehat{h}_n,\ell_0)-\dot{\mathbb{M}}^{(1)}_n(h_0,\ell_0)\\
&=&-\dfrac{1}{n}\sum_{i=1}^n\dfrac{1}{2}\left[(D(\widehat{h}_n,\ell_0)-D(h_0,\ell_0))^2\times\left(\dfrac{\xi_i\delta_i}{D(\widetilde{h}_{1,n},\ell_0)^2}+\dfrac{\xi_i(1-\delta_i)}{(\ell_0-D(\widetilde{h}_{2,n},\ell_0))^2}\right)\right](T_i,X_i)
 \end{eqnarray*}
where $\widetilde{h}_{j,n}(T_i,X_i)$, $j=1,2$ are random values ranging between $\widehat{h}_n(T_i,X_i)$ and $h_0(T_i,X_i)$ almost surely. The assumptions on $D$ allows then to have 
\begin{eqnarray*}
\dfrac{1}{n}\sum_{i=1}^n(\widehat{h}_n-h_0)^2(T_i,X_i)&\lesssim&\left|\mathbb{M}^{(1)}_n(\widehat{h}_n,\ell_0)-Q_\bold{y}m_{\widehat{h}_n,\ell_0}-(\mathbb{M}^{(1)}_n(h_0,\ell_0)-Q_\bold{y}m_{h_0,\ell_0})\right|\\
&+&\left|\dot{\mathbb{M}}^{(1)}_n(\widehat{h}_n,\ell_0)-Q_\bold{y}\dot{m}_{\widehat{h}_n,\ell_0}-(\dot{\mathbb{M}}^{(1)}_n(h_0,\ell_0)-Q_\bold{y}\dot{m}_{h_0,\ell_0})\right|\\
&+&\left|Q_\bold{y}m_{\widehat{h}_n,\ell_0}-Q_\bold{y}m_{h_0,\ell_0}-Q_\bold{y}\dot{m}_{\widehat{h}_n,\ell_0}+Q_\bold{y}\dot{m}_{h_0,\ell_0}\right|.
\end{eqnarray*}
The same argument on the constants $U_1,\dot{U}_1$ and $\sigma_1,\dot{\sigma}_1$ in the step 1.2 ensures that the suprema in (\ref{proof:m1}) and (\ref{proof:mdot1}) are valide over $\mathcal{H}$ and thus
\begin{eqnarray*}
\dfrac{1}{n}\sum_{i=1}^n(\widehat{h}_n-h_0)^2(T_i,X_i)&\lesssim&\left|Q_\bold{y}m_{\widehat{h}_n,\ell_0}-Q_\bold{y}m_{h_0,\ell_0}-Q_\bold{y}\dot{m}_{\widehat{h}_n,\ell_0}+Q_\bold{y}\dot{m}_{h_0,\ell_0}\right|+O(n^{-1/4})\\
&\lesssim& d(\widehat{h}_n,h_0)^2+O(n^{-1/4})
\end{eqnarray*}
which proves the second part of (\ref{proof:speed}) since $q_{n,1}=o(n^{1/4})$ by assumption.\hfill$\Box$

\subsection*{Parameters used in the experiments to generate synthetic data \label{parameters_synthetic_data}}

For each of the two distributions of the survival times, Weibull and Frechet, we consider nine different scenarios with different level of right-censored data (when  $\delta_i=0$) and different level of 
missing delta (when  $\xi_i=0$). The value of the parameters used for the survival times ($a_i$), the censoring mechanism ($b_i$) and the missingness mechnism ($c_i$)  are summarized in Table  \ref{table:parameters_different_scenarios}.

\begin{table}[H]
\centering
\tiny
\begin{tabular}{lccccccccccccccc }
\toprule
Weibull \\
\hline
     &   $\delta=0$ ratio &  $\xi=0$ ratio & \multicolumn{4}{c}{Survival times} & \multicolumn{4}{c}{Censoring mechanism}  & \multicolumn{5}{c}{Missingness mechanism} \\
& & & $a_0$ & $a_1$ & $a_2$ & $a_3$ & $b_0$ & $b_1$ & $b_2$ & $b_3$ & $c_0$ & $c_1$ & $c_2$ & $c_3$ & $c_4$\\
 \hline
 Scenario 1  & 25\%  & 25\% & 2 & 4 & 3 & -0.2 & 0.2 &  -0.1 &  0.2 & 0.1 & 0.1 & 1 & 0.2 & -0.1 & 0.5 \\
 
 Scenario 2  & 25\%  & 50\% 
 
 & 2 & 4 & 3 & -0.2 & 0.2 &  -0.1 &  0.2 & 0.1 & -0.7 & 1 & 0.2 & -0.1 & 0.5 \\
 
 Scenario 3  & 25\%  & 75\% 
 
 & 2 & 4 & 3 & -0.2 & 0.2 &  -0.1 &  0.2 & 0.1 & -1.4 & 1 & 0.2 & -0.1 & 0.5 \\

 Scenario 4  & 50\%  & 25\% 
 
 & 1 & 4 & 3 & -0.2 & 0.9 &  -0.1 & 0.4 & -0.5 & 0.1  & 1 & 0.2 & -0.1 & 0.5\\
 
 Scenario 5  & 50\%  & 50\% & 1 & 4 & 3 & -0.2 & 0.9 &  -0.1 & 0.4 & -0.5 & -0.7 & 1 & 0.2 & -0.1 & 0.5\\
 
 Scenario 6  & 50\%  & 75\% & 1 & 4 & 3 & -0.2 & 0.9 &  -0.1 & 0.4 & -0.5 & -1.3 & 1 & 0.2 & -0.1 & 0.5\\
 Scenario 7  & 75\%  & 25\% & 3 & 4 & 3 & -0.2 & 2.75 & -2 &  0.4 & -0.5 & 0.1 & 1 & 0.2 & -0.1 & 0.5 \\
 Scenario 8  & 75\%  & 50\% & 3 & 4 & 3 & -0.2 & 2.75 & -2 &  0.4 & -0.5 &  -0.6 & 1 & 0.2 & -0.1 & 0.5\\
 Scenario 9  & 75\%  & 75\% & 3 & 4 & 3 & -0.2 & 2.75 & -2 &  0.4 & -0.5 & -1.3 & 1 & 0.2 & -0.1 & 0.5\\
\toprule
Frechet \\
\hline
     &   $\delta=0$ ratio &  $\xi=0$ ratio & \multicolumn{4}{c}{Survival times} & \multicolumn{4}{c}{Censoring mechanism}  & \multicolumn{5}{c}{Missingness mechanism} \\
     & & & $a_0$ & $a_1$ & $a_2$ & $a_3$ & $b_0$ & $b_1$ & $b_2$ & $b_3$ & $c_0$ & $c_1$ & $c_2$ & $c_3$ & $c_4$\\
 \hline
 Scenario 1  & 25\%  & 25\% & 2& 2& 0.2& 1 & 0.2& -0.1& 0.25& 0.1 & 0.1& 1&0.2&-0.1& -0.3\\
 Scenario 2  & 25\%  & 50\% & 2& 2& 0.2& 1 & 0.2& -0.1& 0.25& 0.1  & -0.6& 1&0.2&-0.1& -0.3\\
 Scenario 3  & 25\%  & 75\% & 2& 2& 0.2& 1 & 0.2& -0.1& 0.25& 0.1 & -1.4& 1&0.2&-0.1& -0.3\\
 Scenario 4  & 50\%  & 25\% & 2& 2& 0.2& 1 & 0.9& -0.1& 0.4& -0.5 & 0.1& 1&0.2&-0.1& -0.3 \\
 Scenario 5  & 50\%  & 50\%  & 2& 2& 0.2& 1 & 0.9& -0.1& 0.4& -0.5 & -0.55& 1&0.2&-0.1& -0.3 \\
 Scenario 6  & 50\%  & 75\% & 2& 2& 0.2& 1 & 0.9& -0.1& 0.4& -0.5 & -1.4& 1&0.2&-0.1& -0.3 \\
 Scenario 7  & 75\%  & 25\% & 2& 2& 0.2& 1 & 1.8& -0.1& 0.4& -0.5 & 0.1& 1&0.2&-0.1& -0.3 \\
 Scenario 8  & 75\%  & 50\% & 2& 2& 0.2& 1 & 1.8& -0.1& 0.4& -0.5 & -0.55& 1&0.2&-0.1& -0.3 \\
 Scenario 9  & 75\%  & 75\% & 2& 2& 0.2& 1 & 1.8& -0.1& 0.4& -0.5 & -1.3& 1&0.2&-0.1& -0.3 \\
\bottomrule
\end{tabular}
 \caption{Parameters chosen to generate data with different levels of right-censored data and missing delta with the Weibull and Frechet distributions.}
\label{table:parameters_different_scenarios}
\end{table}

\end{document}